\documentclass[10pt,twocolumn,letterpaper]{article}

\usepackage{iccv}
\makeatletter
\@namedef{ver@everyshi.sty}{}
\makeatother
\usepackage{times}
\usepackage{epsfig}
\usepackage{graphicx}
\usepackage{amsmath}
\usepackage{amssymb}
\usepackage{subfig}
\usepackage{stackengine}
\usepackage{booktabs}
\usepackage{multirow}
\usepackage{tikz}
\usepackage{url}
\usepackage{pgfplots}
\usepgfplotslibrary{units}
\iccvfinalcopy % *** Uncomment this line for the final submission

\def\iccvPaperID{3686} % *** Enter the ICCV Paper ID here

\ificcvfinal\fi
\begin{document}

%%%%%%%%% TITLE
\title{Synthesizing a 4D Spatio-Angular Consistent Light Field from a Single Image}

\author{Andre Ivan$\dagger$\\
{\tt\small andreivan13@gmail.com}
\and
Williem$\ddagger$\\
{\tt\small williem@binus.edu}
\and
In Kyu Park$\dagger$\\
{\tt\small pik@inha.ac.kr}
\and
$\dagger$Dept. of Information and Communication Eng., Inha University, Incheon 22212, Korea\\
\and
$\ddagger$School of Computer Science, Bina Nusantara University, Jakarta, Indonesia 11480\\
}

\maketitle

\begin{abstract}
Synthesizing a densely sampled light field from a single image is highly beneficial for many applications. The conventional method reconstructs a depth map and relies on physical-based rendering and a secondary network to improve the synthesized novel views. Simple pixel-based loss also limits the network by making it rely on pixel intensity cue rather than geometric reasoning. In this study, we show that a different geometric representation, namely, appearance flow, can be used to synthesize a light field from a single image robustly and directly. A single end-to-end deep neural network that does not require a physical-based approach nor a post-processing subnetwork is proposed. Two novel loss functions based on known light field domain knowledge are presented to enable the network to preserve the spatio-angular consistency between sub-aperture images effectively. Experimental results show that the proposed model successfully synthesizes dense light fields and qualitatively and quantitatively outperforms the previous model . The method can be generalized to arbitrary scenes, rather than focusing on a particular class of object. The synthesized light field can be used for various applications, such as depth estimation and refocusing.
\end{abstract}

%%%%%%%%% BODY TEXT
%-------------------------------------------------------------------------
\begin{figure}[t]
\begin{center}
	{\includegraphics[width=0.24\linewidth]{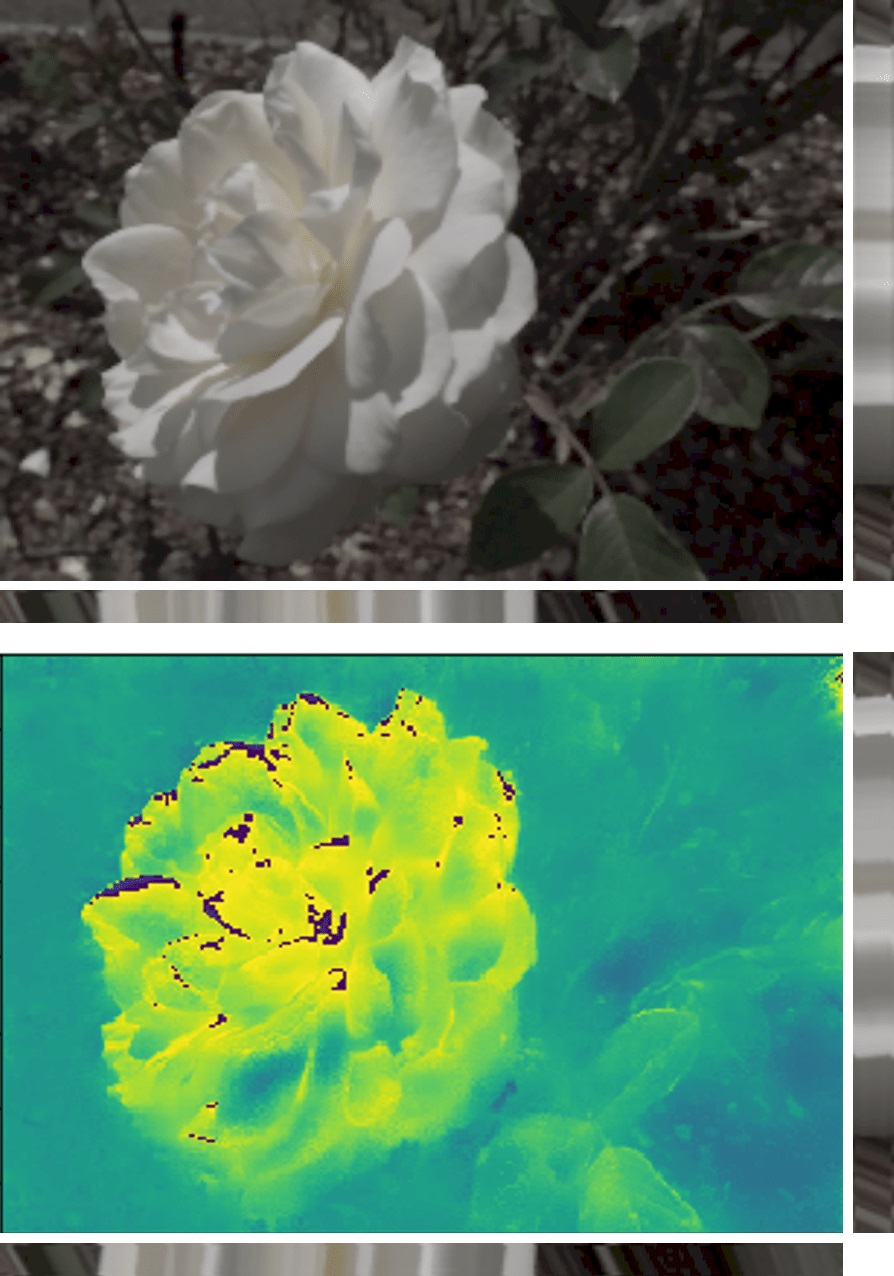}}~%
  	{\includegraphics[width=0.5\linewidth]{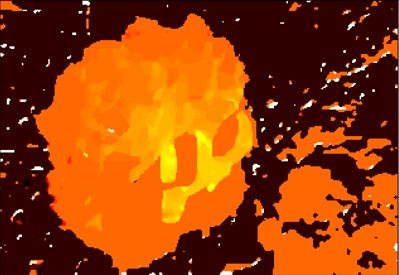}}~%
	{\includegraphics[width=0.25\linewidth]{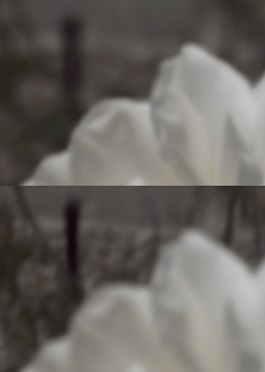}}~%

	\vspace{-3.0mm}
	\subfloat[]{\includegraphics[width=0.24\linewidth]{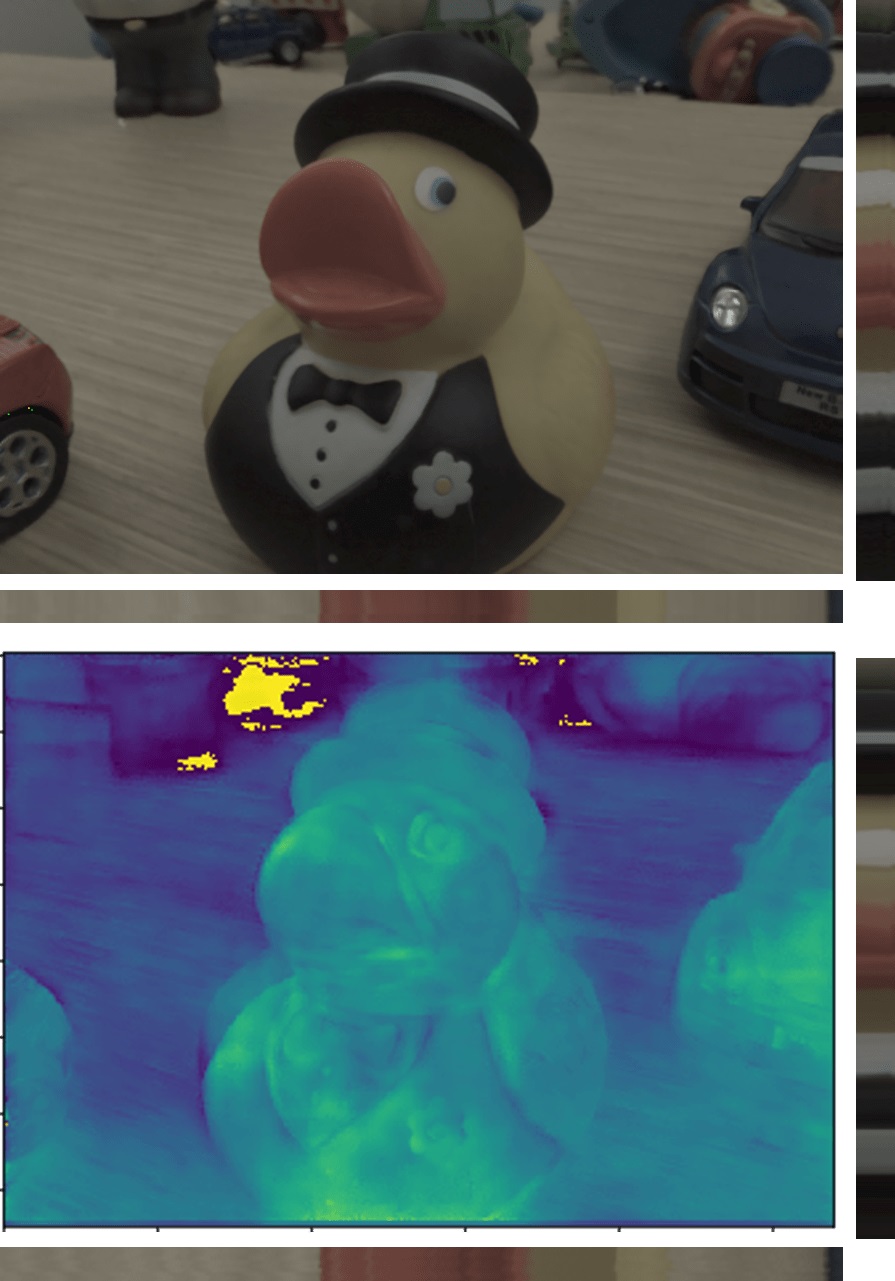}}~%
    \subfloat[]{\includegraphics[width=0.5\linewidth]{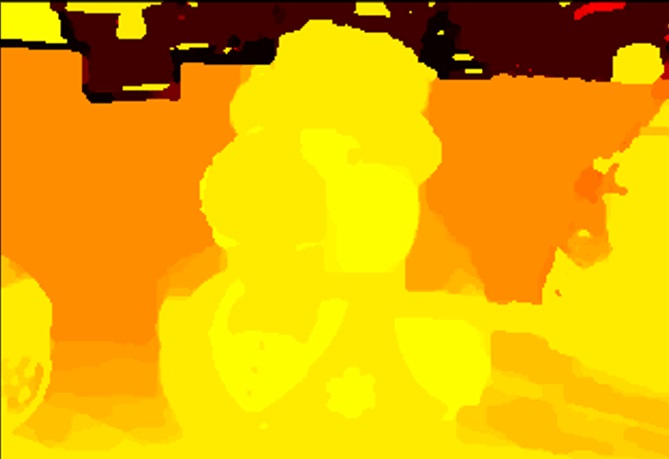}}~%
	\subfloat[]{\includegraphics[width=0.25\linewidth]{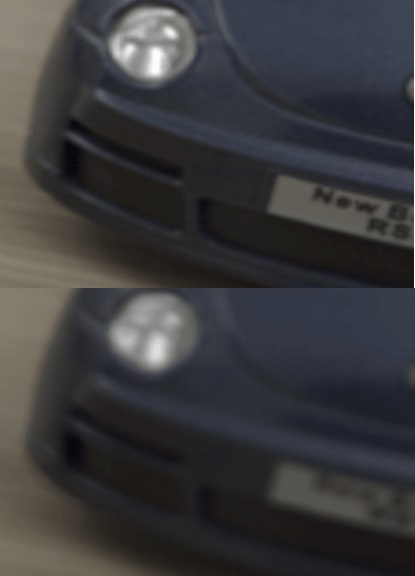}}~%	
    \vspace{-4.5mm}
\end{center}
\caption{Applications of the synthesized light field. (a) Input image and estimated appearance flow with ground truth and estimated EPI, respectively; (b) Depth image; (c) Patch of refocused image at foreground and background.}
    \label{fig:Intro_Fig}
    \vspace{-3.0mm}
\end{figure}
%------------------------------------------------------------------------
\section{Introduction}
Light fields have attracted considerable interest from computer vision and graphic communities due to their capability to capture multiple light rays from various directions.
Recent studies on light fields utilized densely sampled light fields captured by light field cameras.
Many applications, such as depth estimation~\cite{Schilling_CVPR18,Shin_CVPR18,Williem_CVPR16,Williem_PAMI17}, refocusing~\cite{Ng_CSTR05}, and 3D reconstruction~\cite{Johannsen_ACCV16,Vianello_CVPR18}, exploit the rich information of light fields.
Promoting a single image into a densely sampled light field can also elevate existing AR/VR experiences.

At present, a light field image is captured using either plenoptic cameras~\cite{Raytrix_2013} or camera arrays~\cite{Wilburn_TOG05}.
However, the absence of the only available consumer light field camera, {\em i.e.} Lytro, has created a gap between consumers and light field experiences.
We focus on filling this gap so that end users can experience the advantages of light field imaging.
The idea is to synthesize a light field using only a single image, which can be easily obtained in the real world.
Synthesizing a 4D light field is a severely ill-posed problem, but the impact of such work is considerably significant.
The geometry information from a single image should be inferred and used to synthesize the surrounding angular images.

Light field synthesis has attracted considerable attention in recent years~\cite{Kalantari_TOG16,Srinivasan_ICCV17,Wang_ECCV18,Wu_CVPR17,Yeung_ECCV18}.
Previous approaches can be grouped into two based on the input type, namely, single and multi-view inputs.
The multi-view input utilizes multiple images from specific viewpoints to infer the geometric clue and use it to synthesize the light field.
However, only a few consumer cameras can simultaneously capture multi-view images, thus making the approach unsuitable for general use.

Existing methods involving a single input utilize a depth image-based rendering (DIBR) technique to synthesize the light field~\cite{Srinivasan_ICCV17}.
The work is inspired by previous view synthesis techniques using geometry estimation~\cite{Flynn_CVPR16,Garg_ECCV16,Godard_CVPR17,Xie_ECCV16}.
However, ~\cite{Srinivasan_ICCV17} is highly dependent on the estimated depth quality and physical-based depth warping to synthesize angular images.
The depth-based approach also suffers from a non-Lambertian surface and difficulty in reconstructing the occlusion region.

In this study, we develop a novel deep neural network for light field synthesis that utilizes the appearance flow to synthesize novel views.
We also introduce a spatio-angular consistent loss function that helps the network learn robustly and efficiently.
The loss function consists of global and local losses.
A post-processing method is applied to improve the quality of complex scenes.
The experimental results show that the proposed method outperforms the conventional single image-based approach in qualitative and quantitative evaluations.
Figure~\ref{fig:Intro_Fig} shows the results of the proposed method.

The key contributions of this study are summarized as follows.
\begin{itemize}
\itemsep0.01em
   \item Robust alternative geometric representation to synthesize light field without using physical-based approach.
   \item Novel spatio-angular consistent loss that imposes geometric reasoning to the network.
   \item Capability to be generalized to arbitrary scenes rather than a specific class of object compared with the previous approach.
   \item Single and fast end-to-end full learning based deep neural network model.
\end{itemize}
%------------------------------------------------------------------------
%------------------------------------------------------------------------
\begin{figure*}[t]
\begin{center}
    {\includegraphics[width=1.0\linewidth]{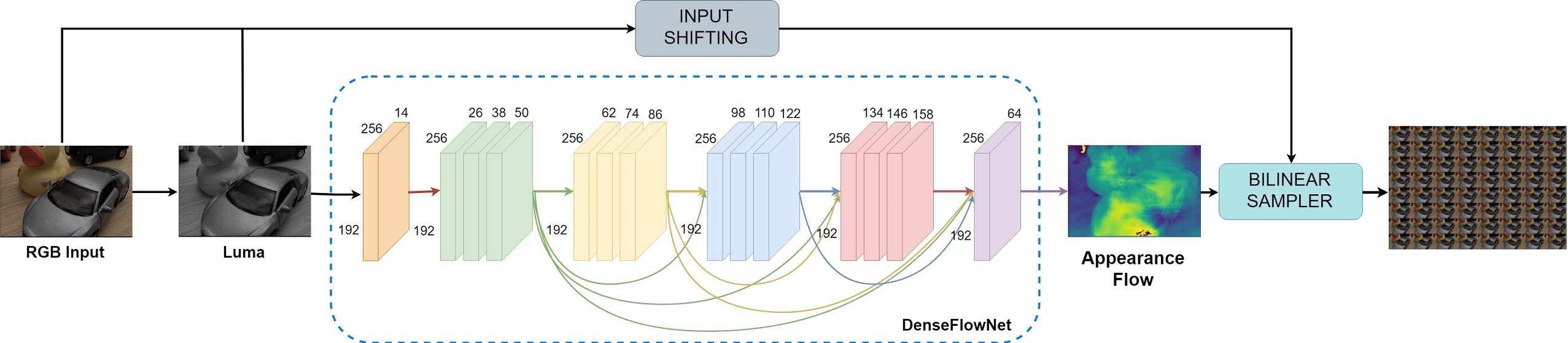}}
\end{center}
    \vspace{-3.0mm}
\caption{Proposed network structure. The image is pre-convolved by a single convolution layer. The following 4 blocks are densely connected block each with 3 dilated convolution layers. The last convolution layer estimates $(u\times v \times 2)$ flows in $x$ and $y$ direction. The RGB and the luminance ($Y $) image are shifted producing $Y $ and RGB light field, where the former is used for loss computation and the later as output of the network.}
    \label{fig:Network_Structure}
    \vspace{-1.5mm}
\end{figure*}
%------------------------------------------------------------------------
\section{Related Works}
Learning based light field synthesis has been investigated by many researchers in the past few years~\cite{Kalantari_TOG16,Srinivasan_ICCV17,Wang_ECCV18,Wu_CVPR17,Yeung_ECCV18,Ruan_EG18}.
On the basis of the number of input images, related approaches are divided into multi-view image~\cite{Kalantari_TOG16,Wu_CVPR17,Wang_ECCV18,Yeung_ECCV18} and single image~\cite{Srinivasan_ICCV17,Ruan_EG18} based methods.
Multi-view image-based light field synthesis is impractical due to its specific input pattern.
Meanwhile a single image-based light field synthesis is severely ill-posed although it is the most practical approach for light field synthesis.

\vspace{-3.0mm}
\paragraph{Sparse-input Light Field Synthesis}
Wanner and Goldluecke~\cite{Wanner_PAMI14} introduced a light field super resolution framework adopting estimated depth information and variational optimization to fill missing views by using a sparse light field image.
Phase-based light field synthesis from a micro-baseline stereo pair was proposed by Zhang~\etal~\cite{Zhang_CVPR15}.
Those studies were rooted on traditional approaches that use complex processing and various optimization approaches.
Meanwhile, learning-based view synthesis produces better results by using an end-to-end training strategy.
Zhou~\etal~\cite{Zhou_ECCV16} proposed a new geometric representation called appearance flow to synthesize image with a novel view.
However, the proposed representation did not generalize well to a scene with multiple objects and non-homogenous background.
Flynn~\etal~\cite{Flynn_CVPR16} introduced a deep learning approach to synthesize a novel view with a wide baseline using images with their pose information.
Zhou~\etal~\cite{Zhou_TOG18} presented a novel geometric representation called multi-plane images to synthesize a horizontal light field from a narrow baseline camera.
However, these approaches require camera pose information to synthesize the novel view which is not commonly available in the real world.

Kalantari~\etal~\cite{Kalantari_TOG16} introduced the first learning based light field synthesis solution.
They utilized four corner images to synthesize a 4D light field using a depth estimation and color refinement approach.
The inputs to the depth estimation network were mean and variance images, as inspired by the depth estimation work of~\cite{Tao_ICCV13}.
Wu~\etal~\cite{Wu_CVPR17} utilized an epipolar plane image~(EPI) obtained from sparse input images and synthesized an up-sampled EPI through a specially designed blur kernel.
Wang~\etal~\cite{Wang_ECCV18} employed a pseudo 4DCNN represented as 2D strided convolution and 3DCNN, where the light field image was synthesized in a step-by-step manner by fixing each angular resolution.
Yeung~\etal~\cite{Yeung_ECCV18} applied a high dimensional convolutional kernel to infer spatial and angular information from sparse input images.
In summary, sparse input image-based light field synthesis focuses on synthesizing in-between views and could be regarded as solving an interpolation problem.
The specific input sampling pattern hinders its practical usage.

\vspace{-3.0mm}
\paragraph{Single-input Light Field Synthesis}
Srinivasan~\etal~\cite{Srinivasan_ICCV17} introduced the first solution to solve light field synthesis from a single image.
They proposed a single image based depth estimation to obtain the approximate geometry of a scene.
Then, the estimated depth is utilized to synthesize a novel view using the DIBR approach.
However, their method is constrained to a non-Lambertian scene and highly dependent on pixel intensity.
Ruan~\etal~\cite{Ruan_EG18} used a Wasserstein GAN to synthesize a light field image.
The framework relies on GAN capability guided by perceptual and EPI losses to synthesize novel views.
In this study, we focus on solving the problem of single-image light field synthesis.
Contrary to~\cite{Srinivasan_ICCV17}, we propose an alternative geometric representation to synthesize a light field image.
Furthermore, we present a light field based loss function that enables the network to learn geometric reasoning.
%------------------------------------------------------------------------
\section{Proposed Method}
\subsection{Light Field Synthesis Formulation}
This study aims to synthesize a 4D light field ${L}(x,y,u,v)$ given a single image that serves as the central sub-aperture image~(SAI) ${L}(x,y,0,0)$.
We follow the two-plane parametrization light field ${L}(x,y,u,v)$, introduced by~\cite{Levoy_CGI96}, where $(x,y)$ and $(u,v)$ are the coordinates in spatial and angular planes, respectively.
In general, the light field synthesis problem is formulated as DIBR problem, which is described as follows:
\begin{equation}
{L}(x,y,u,v) = {L}(x + d(u),y + d(v),0,0),
\end{equation}
where $d(u)$ and $d(v)$ are the disparity in $x$ and $y$ directions, respectively.
Disparity depends on the depth information of the central image and the novel angular coordinate $(u, v)$.
We address the light field synthesis problem by using an approximation function $f(\cdot)$ represented as a deep convolutional neural network rather than relying on depth information, as described in
\begin{equation}
{L}(x,y,u,v) = f({L}(x,y,0,0)).
\end{equation}
Function $f$ solves a highly ill-posed problem.

An appearance flow is estimated to extrapolate the central view to each SAI in the 4D light field.
Appearance flow represents 2D  coordinate vectors specifying where to reuse or copy pixels to reconstruct the novel view.
Considering that the ground truth light field appearance flow is difficult and expensive to obtain, the proposed network is designed to estimate appearance flow in an unsupervised manner.
Appearance flow~\cite{Zhou_ECCV16} is accompanied with few blurs, preserves color identities, and removes the dependency on the physical-based approach to synthesize novel views.

To make the problem tractable, function $f$ is decomposed into three sub-problems, namely, appearance flow estimation for each viewpoint $(u,v)$, image shifting with respect to the central view, and novel view extrapolation:
Each sub-problem can be defined as follows.
\begin{equation}
{L}_f(x,y,u,v) = \mathcal{F}({L}(x,y,0,0))
\end{equation}
\begin{equation}
{L}_s(x,y,u,v) = \mathcal{S}({L}(x,y,0,0), \nabla(u,v))
\end{equation}
\begin{equation}
\hat{L}(x,y,u,v) = \mathcal{W}({L}_s(x,y,u,v), {L}_f(x,y,u,v)),
\end{equation}
where $\mathcal{F}$ is the proposed {\it DenseFlowNet} that estimates appearance flow ${L}_f$ for each novel angular view.
$\mathcal{S}$ performs angular shifting to the position $\nabla(u,v)$ of novel views.
$\mathcal{W}$ is the warping function for shifted image ${L}_s(x,y,u,v)$ using its corresponding appearance flow ${L}_f(x,y,u,v)$.
Image warping is performed using a bilinear sampler module~\cite{Jaderberg_NIPS15} to produce  light field image $\hat{L}(x,y,u,v)$.

The proposed objective function is defined as
\begin{equation}
\begin{aligned}
\underset{\theta}{\text{min}}
    \sum[
            \lambda{_g}L_{g}(\theta)+
            \lambda{_{e}}L_{e}(\theta) +
            \lambda{_{tv}}\psi_{tv}(\theta)
            ],
\end{aligned}
\end{equation}
where $\theta$ denotes the deep neural network parameters.
The problem formulation and objective function enable the network to estimate an appearance flow for each SAI $(u,v)$ in an unsupervised manner through the supervision of synthesized pixels.
The common pixel-wise loss is not utilized in the proposed method because it does not enforce the geometry constraint to the network.
Instead, we rely on the known light field domain knowledge and design two geometrically constrained losses, {\em i.e.} global $L_{g}(\theta)$ and local $L_{e}(\theta)$ light field losses.
Both loss functions are useful in preserving the spatio-angular consistency between light field SAIs.
In addition, we propose a regularization loss $\psi_{tv}(\theta)$ to the estimated appearance flow to reduce the jittering artifacts.
Each loss is weighted using the global loss weight $\lambda{_g}$ and the local loss weight $\lambda{_e}$.
$\psi_{tv}$ is the flow regularization weight.
The proposed network consists of mostly convolution operations without physical-based computation, resulting in low computational complexity.
The overall network structure is shown in Figure~\ref{fig:Network_Structure}.
The details of each sub-problem are explained in the following sections.
%-------------------------------------------------------------------------
\begin{figure}[t]
\begin{center}
    \stackunder[5pt]{\includegraphics[width=0.49\linewidth]{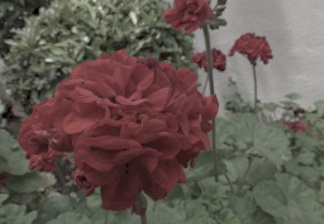}}{Reference image}%
    \stackunder[5pt]{\includegraphics[width=0.49\linewidth]{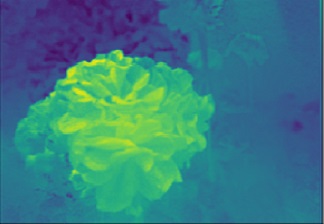}}{Appearance flow}%
\end{center}
\vspace{-3.5mm}
\caption{Visualization of the estimated appearance flow using {\em DenseFlowNet}. The estimated appearance flow successfully identifies and separates the object of interest in the scene.}
    \label{fig:Flow_Visualization}
\vspace{-2.0mm}
\end{figure}
%-------------------------------------------------------------------------
\subsection{Appearance Flow Estimation}
To estimate an accurate and dense appearance flow, {\em DenseFlowNet} is designed to extract all important representations in the image.
The network consists of dense skip connections that connect each layer with the following layers.
In the proposed network, we use four dense blocks in which each block has three convolution layers followed by the ReLU activation function~\cite{Nair_ICML10}.
Batch normalization and average pooling in the transition layer are removed.
Instead, we use dilated convolution to allow the network to have access to the entire input image without reducing its spatial resolution.
An extra convolution block at the final layer estimates the flow for each novel view.
The final convolution layer does not have an activation function.
Additional details about the network structure, \ie kernel size and dilation rate, are available in the supplementary material.

In particular, the proposed flow estimation network produces coordinate vectors $(x,y)$ to sample from ${L}_s(x,y,u,v)$, which is used by the bilinear sampler to synthesize the novel view $\hat{L}(x,y,u,v)$.
The appearance flow is visualized in Figure~\ref{fig:Flow_Visualization}.
The brighter pixel indicates the pixel have bigger displacement.
The input to the network is the luminance ($Y$) component which is used to compute the loss.
The luminance component is widely used in super-resolution~\cite{Dong_PAMI16} and light field synthesis~\cite{Yeung_ECCV18}.
We learn $\mathcal{F}$ to estimate the appearance flow for all novel views by minimizing the loss between the extrapolated novel views $\hat{L}(x,y,u,v)$ and ground truth light field ${L}(x,y,u,v)$.

\subsection{Image Shifting}
Appearance flow represents the coordinates in $x$ and $y$ directions only.
Therefore, the information in $z$ (depth) direction is lacking.
This condition leads to a poor result and contains unpleasing artifacts when compared with DIBR approaches~\cite{Kalantari_TOG16,Srinivasan_ICCV17} that have pre-defined geometrical constraints.

To alleviate this problem, we propose an input shifting technique $\mathcal{S}$, for light field synthesis with respect to the novel view coordinate.
This technique is inspired by the work of Xie~\etal~\cite{Xie_ECCV16}, in which the input image is shifted to guide the network in synthesizing the corresponding stereo image.
We extend it to three directions ({\em i.e.} horizontal, vertical, and diagonal).
The shifting operation can be formally written as follows:
\begin{equation}
{L}_s(x,y,u,v) = {L}(x-(\eta \times \Delta u),y-(\eta \times \Delta v),0,0),
\end{equation}
where $\eta$ is the constant angular shift value in horizontal and vertical directions.
$\Delta u$ and $\Delta v$ are the angular distances between novel and central views.
Considering the redundancy in light field SAI, we can partially imitate how a pixel shifts in each angular position and utilize this to provide better intialization.
The $\eta$ value is fixed based on the disparity between SAI in the target light field.
The value of $\eta$ corresponds to the width of the light field baseline.

\begin{figure}[t]
\vspace{-3.0mm}
\centering
    \subfloat[Reference image]{\includegraphics[width=0.6\linewidth]{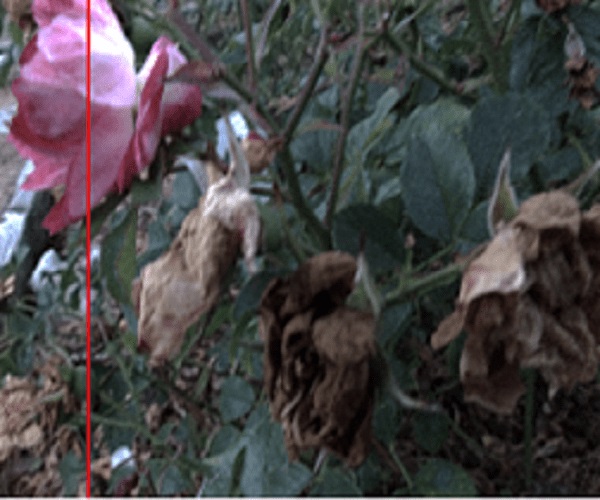}}~%
    \subfloat[]{\includegraphics[width=0.2\linewidth]{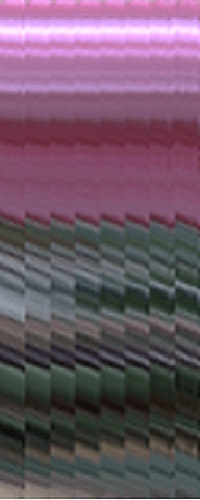}}~%
    \subfloat[]{\includegraphics[width=0.2\linewidth]{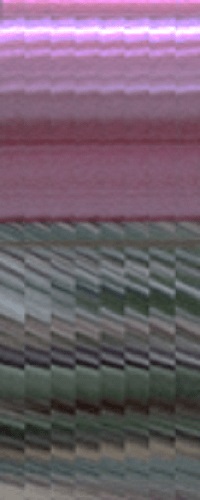}}~%
%\vspace{-1mm}
\caption{Image shifting visualization. Reference image (a) and stacked EPI sliced for each $v$ (b)(c). (b) Stacked EPI from the shifted input; (c) Stacked EPI from the ground truth light field. Parts of (b) resemble (c) especially in the background region.}
    \label{fig:Image_Shifting_Analysis}
\vspace{-3.0mm}
\end{figure}

Figure~\ref{fig:Image_Shifting_Analysis} shows an image shifting result, in which several regions in EPI exhibit similar slopes to the ground truth light field.
Considering that image shifting does not consider parallax, the network is required to estimate the flow that fixes the incorrectly shifted pixels.
Compared with accurate flow estimation of the entire pixels in the image, the network is only required to fix a few regions in the image.
Furthermore, image shifting reduces the range of the estimated appearance flow in the outer angular resolution.
%-------------------------------------------------------------------------
%-------------------------------------------------------------------------
%-------------------------------------------------------------------------
\begin{figure}[t]
\begin{center}
    {\includegraphics[width=1.0\linewidth]{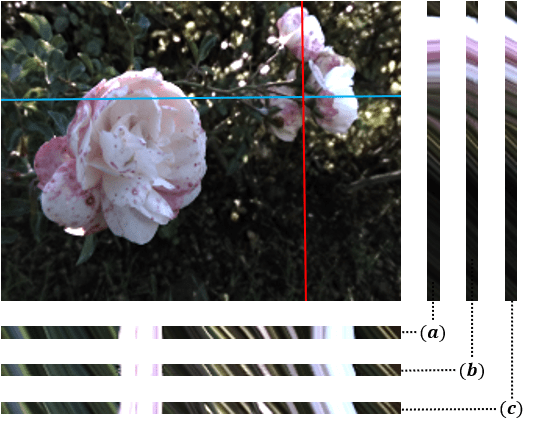}}
    \vspace{-7.0mm}
\end{center}
\caption{Light field based loss result. (a) Result with L1 loss; (b) Result with local loss; (c) Ground truth. From the EPI, it can be seen that (a) perform an inverted shift. The flower in the back should have steeper slope alongside the background pixels in EPI. L1 loss encourage the network to relies on intensity cue and assume objects with similar color relies at the same depth.}
    \label{fig:EPI_Loss}
		\vspace{-2.0mm}
\end{figure}
%-------------------------------------------------------------------------
\subsection{Light Field Loss}
Although image shifting alleviates the amount of network training burden, a good objective function for learning the relation between angular views should be developed.
Simple pixel-wise loss, which is commonly used by conventional approaches, cannot provide proper geometric reasoning to the network.
It encourage the network to look at dominant pixel color individually instead of understanding the whole scene.
We propose a novel 4D light field loss, which is formulated as
\begin{equation}
\begin{aligned}
L_g(\theta) = |M(\hat{L}^\theta(x,y,u,v)) - M({L}(x,y,u,v))| +\\  |V(\hat{L}^\theta(x,y,u,v)) - V({L}(x,y,u,v))|,
\end{aligned}
\end{equation}
where $M(L)$ and $V(L)$ are the mean and variance of a light field computed for all pixel in the SAIs as follows.
\begin{flalign}
M(L) = & \frac{1}{N}\displaystyle\sum_{n}L(x,y,u,v) \\
V(L) = & \sqrt{\frac{1}{N-1}\displaystyle\sum_{n}(L(x,y,u,v)- M(L))^2}
\end{flalign}

Computing the light field mean is equivalent to obtaining the refocus image at zero disparity.
The refocused image correlates to the depth of the light field image.
Therefore, the synthesized light field depth can be explicitly evaluated in an efficient way.
Meanwhile, variance captures the difference between SAIs and helps the network learn the occlusion and edge region.
This is known from the light field depth estimation work of~\cite{Tao_ICCV13,Williem_CVPR16,Williem_PAMI17}, who utilized two representations to compute defocus and correspondence responses.
Kalantari~\etal~\cite{Kalantari_TOG16} also employed the mean and variance image as the input to their depth estimation network.
Here, we show that mean and variance image can be used as loss function to help the network learn the light field geometry.

Although 4D global loss captures geometric information globally, the network should learn the local geometric relation between SAIs in a refined manner.
The idea is to help the network explicitly understand the shifting of pixels the horizontal and vertical directions in the angular domain.
We compute the mean and variance for each row and column in the 4D light field, similar to the direction of EPI sliced conventionally.
The losses at each row and column are accumulated to obtain the final local loss.
The process can be formulated as
%\begin{equation}
\begin{flalign}
& L_e(\theta)= \\\nonumber
    & \displaystyle\sum_{i,j=1}^{I,J} |M(\hat{L}^\theta(x,y,u_{ij},v_{ij})) - M(L(x,y,u_{ij},v_{ij}))| +\\\nonumber
    & \displaystyle\sum_{i,j=1}^{I,J} |V(\hat{L}^\theta(x,y,u_{ij},v_{ij})) - V(L(x,y,u_{ij},v_{ij}))|.
\end{flalign}
%\end{equation}

The objective function encourages the network to learn the fine geometric relation between SAIs by supervising a particular direction.
Figure~\ref{fig:EPI_Loss} shows an example to prove the effect of local light field loss.
The proposed loss overcome the simple pixel-based loss limitation.
%-------------------------------------------------------------------------
\begin{figure}[t]
\vspace{-3.0mm}
\centering
    \subfloat[]{\includegraphics[width=0.5\linewidth]{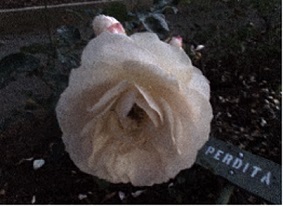}}~%
    \subfloat[]{\includegraphics[width=0.5\linewidth]{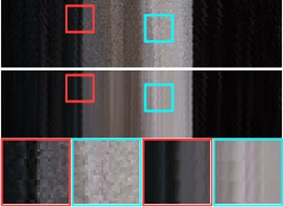}}~%
    \vspace{-2.0mm}
\caption{Denoising capability. (a) Noisy input; (b) First row is the ground truth of noisy EPI and second row is the synthesized EPI. Third row shows the close up view of noisy EPI and synthesized EPI, respectively.}
    \label{fig:Total_Variation}
    \vspace{-2.0mm}
\end{figure}
%-------------------------------------------------------------------------
\subsection{Loss Regularization}
An inconsistent appearance flow might appear and cause artifacts between SAIs.
This problem is expected because appearance flow is estimated from a single image in an unsupervised manner.
Ground truth light field appearance flow is required to fix this issue.
However, it is expensive and difficult to obtain, especially in the real-world scene.
An alternative approach is to use the conventional flow estimation method into the ground truth light field and compare it with the estimated flow.
However, this approach is tedious and increases framework complexity.
Thus, we present a strategy to remedy inconsistent and incorrect flow by incorporating a regularization term into the loss function.

Artifacts mostly occur in the region with fine edges due to the incorrect representation extracted by the network.
Total variation regularization is applied to estimate the flow and thus handle artifacts.
Total variation is commonly used for noise removal in image processing.
The idea is to smoothen noisy flow while keeping edge information.
We show that total variation can also be applied to suppress noise in the flow.
$L_2$ minimization is performed on the gradient of the estimated flow is performed.
Therefore, regularization is performed separately for each coordinate $(x, y)$ and then combined, as described in
\begin{equation}
\psi_{tv}(\theta)= |\nabla_{x} \hat{L}^\theta(x,y,u,v)|_2 + |\nabla_{y} \hat{L}^\theta(x,y,u,v)|_2.
\end{equation}

A small weight is applied to control the regularization effect.
We want the network to estimate diverse and sparse flow for complex scenes while suppressing inconsistent flow.
Another advantage of flow regularization is that the network can handle noisy input and synthesize a denoised light field as shown in Figure~\ref{fig:Total_Variation}.
%------------------------------------------------------------------------
\begin{figure}
\begin{center}
	{\includegraphics[width=\linewidth]{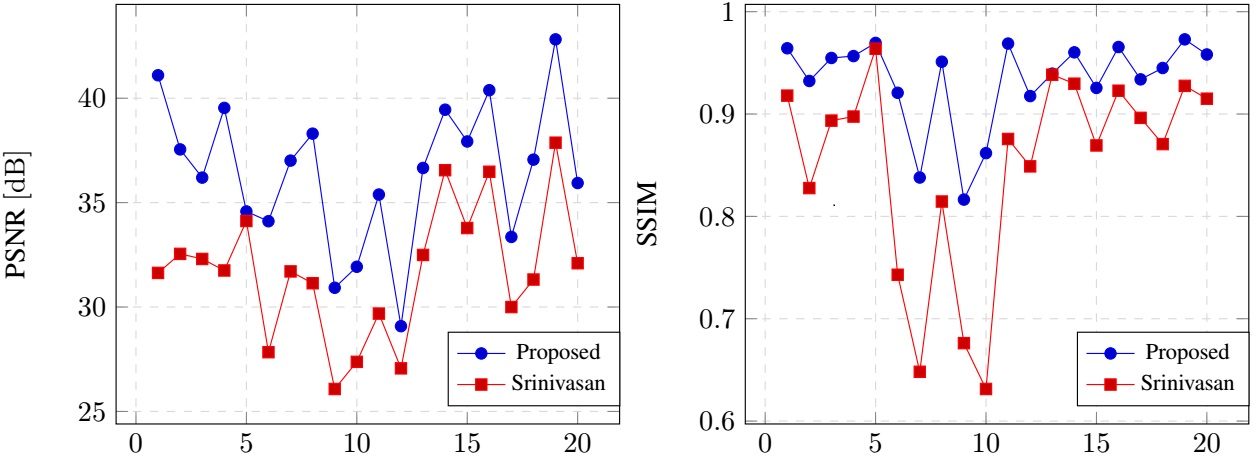}}~%
    \vspace{-2.0mm}
\end{center}
\caption{Quantitative comparison from 20 samples of \textit{Flower} test set.}
    \label{fig:Quantitative_Flower}
    \vspace{-2.0mm}
\end{figure}
%------------------------------------------------------------------------
\begin{figure*}
\begin{center}
    
    {\includegraphics[width=0.33\linewidth]{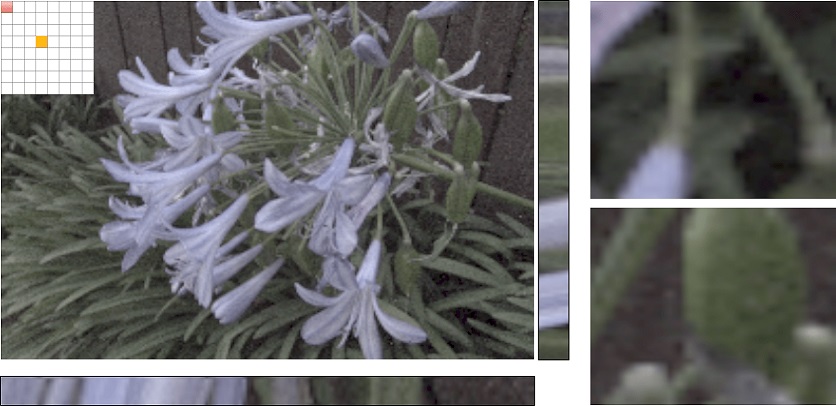}}~%
    {\includegraphics[width=0.33\linewidth]{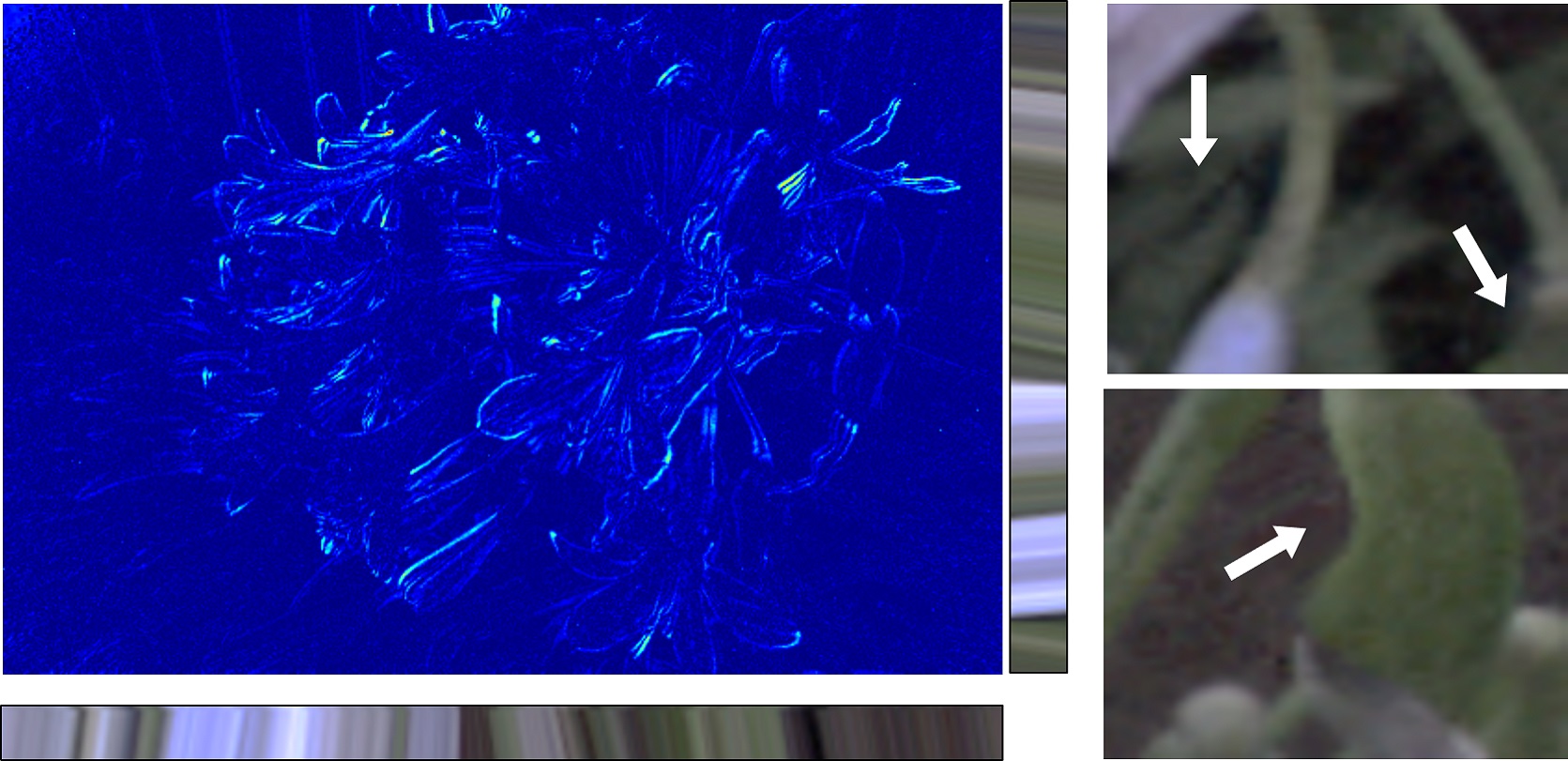}}~%
    {\includegraphics[width=0.33\linewidth]{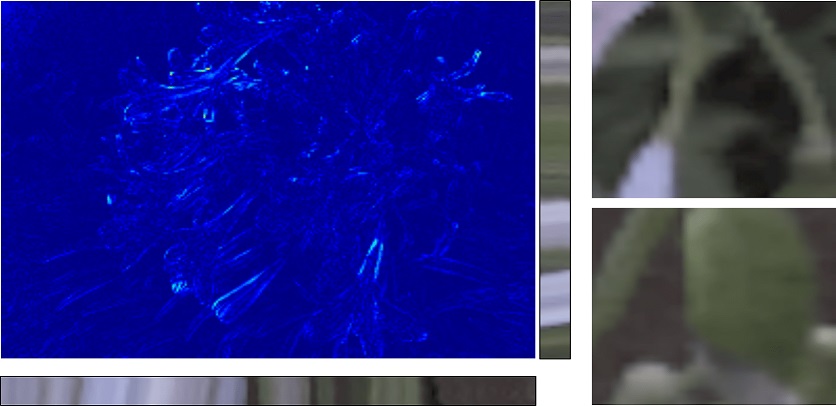}}~%
    
    \vspace{1.0mm}
    {\includegraphics[width=0.33\linewidth]{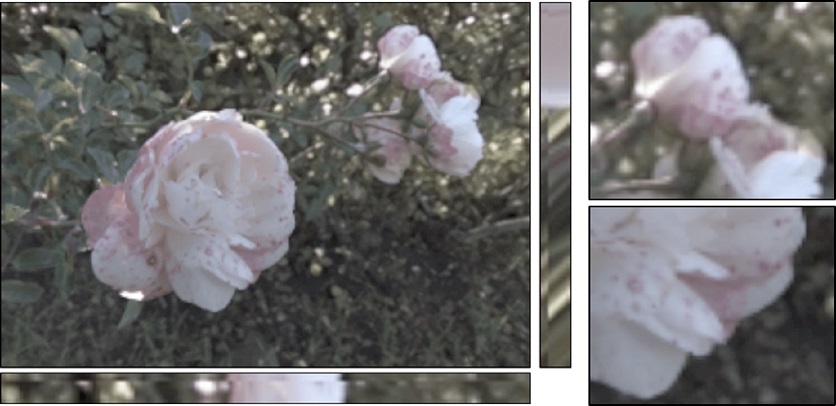}}~%
    {\includegraphics[width=0.33\linewidth]{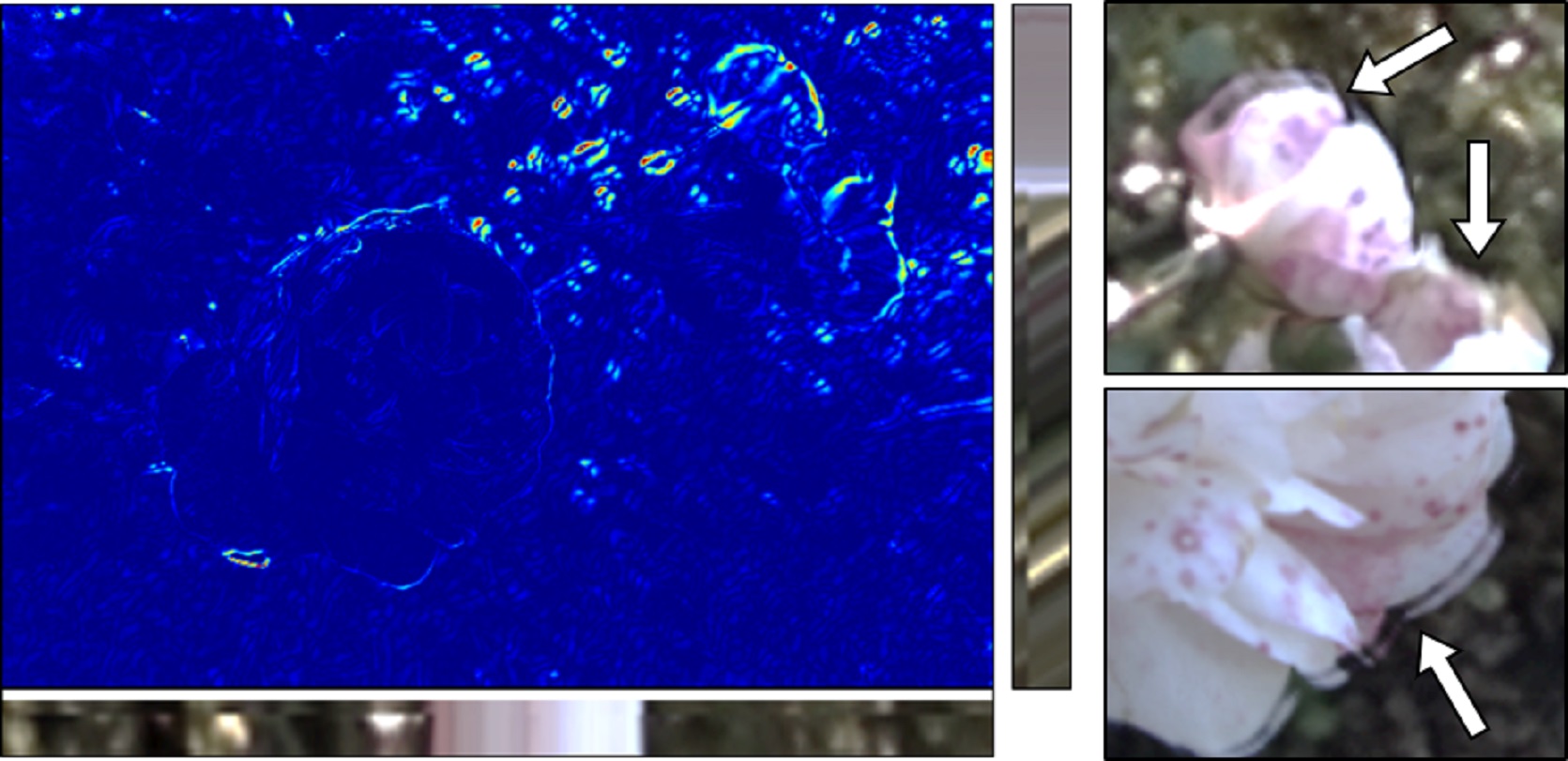}}~%
    {\includegraphics[width=0.33\linewidth]{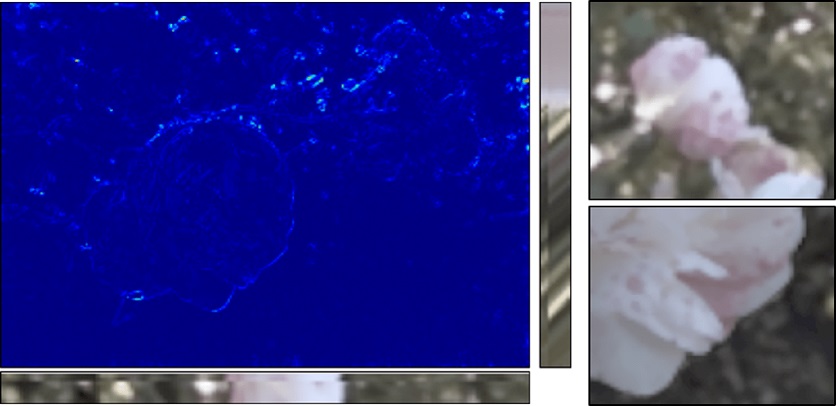}}~%
    
    \vspace{1.0mm}
    {\includegraphics[width=0.33\linewidth]{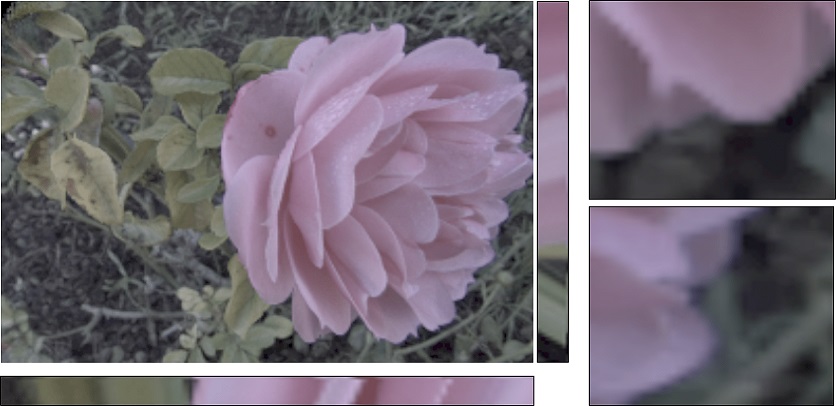}}~%
    {\includegraphics[width=0.33\linewidth]{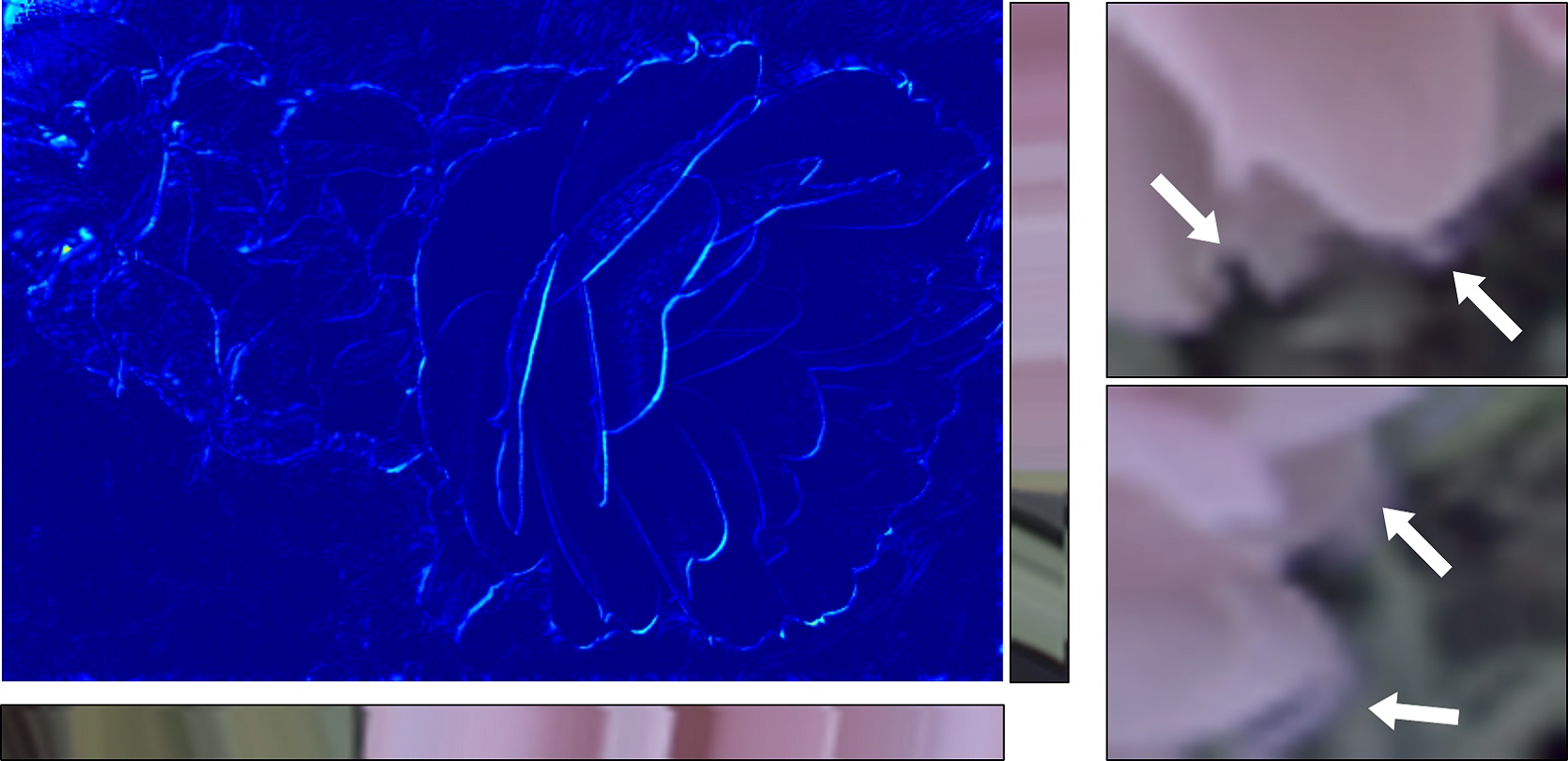}}~%
    {\includegraphics[width=0.33\linewidth]{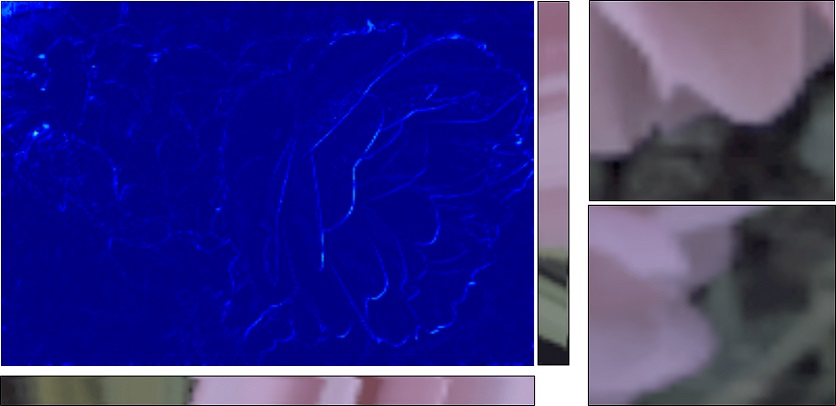}}~%
    
    \vspace{1.0mm}
    {\includegraphics[width=0.33\linewidth]{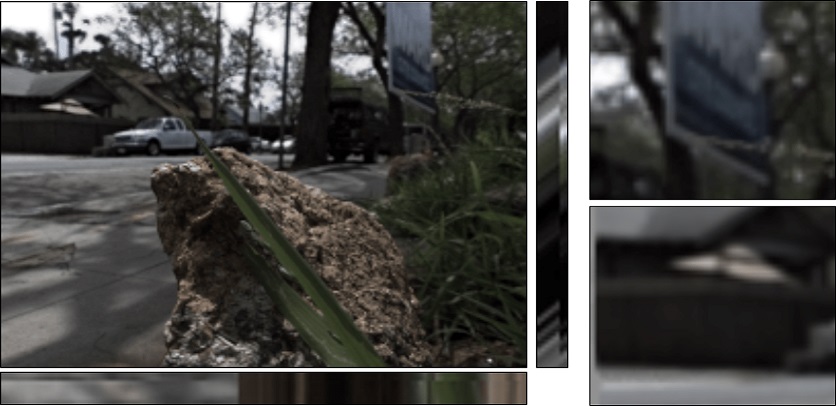}}~%
    {\includegraphics[width=0.33\linewidth]{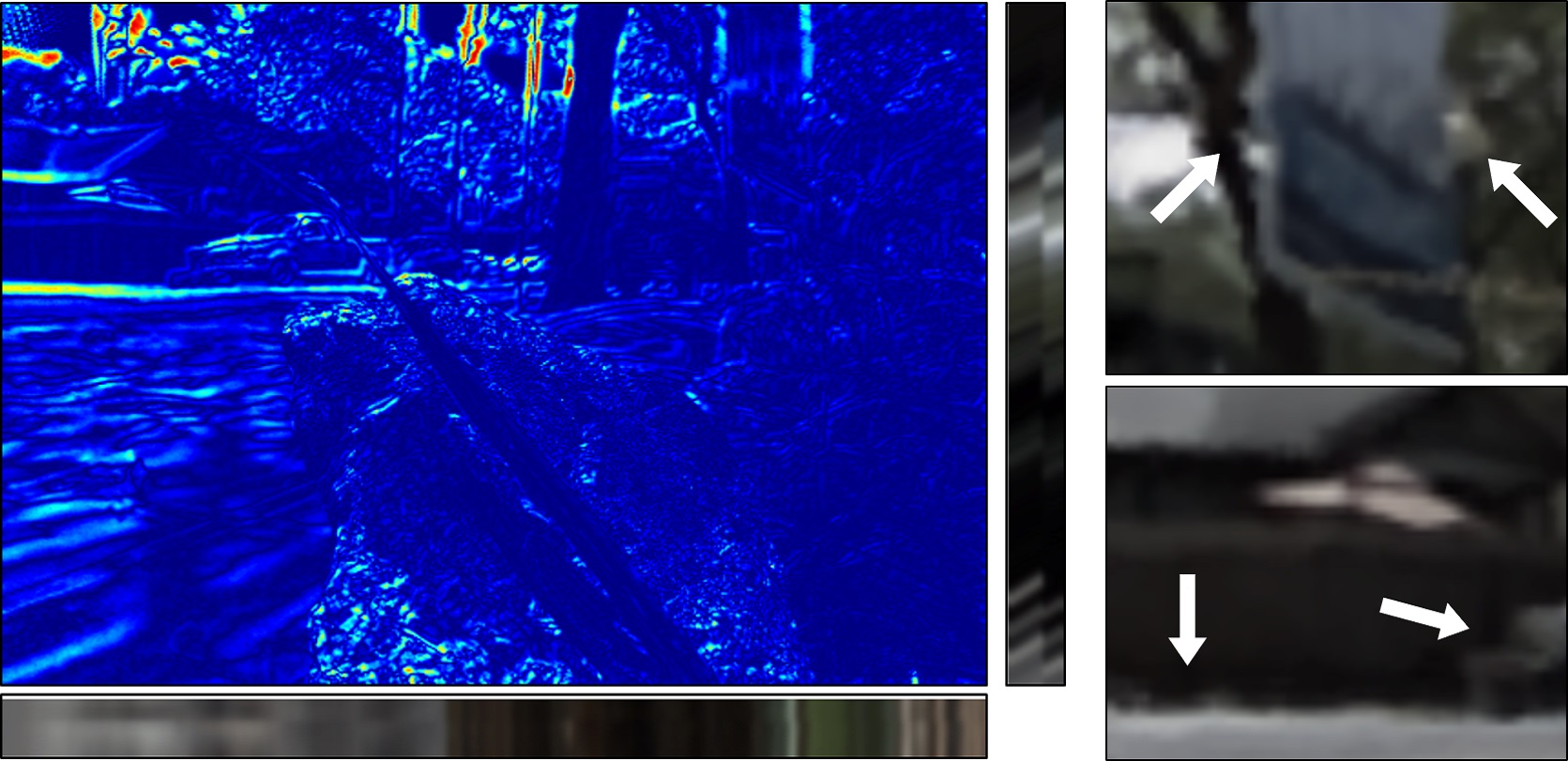}}~%
    {\includegraphics[width=0.33\linewidth]{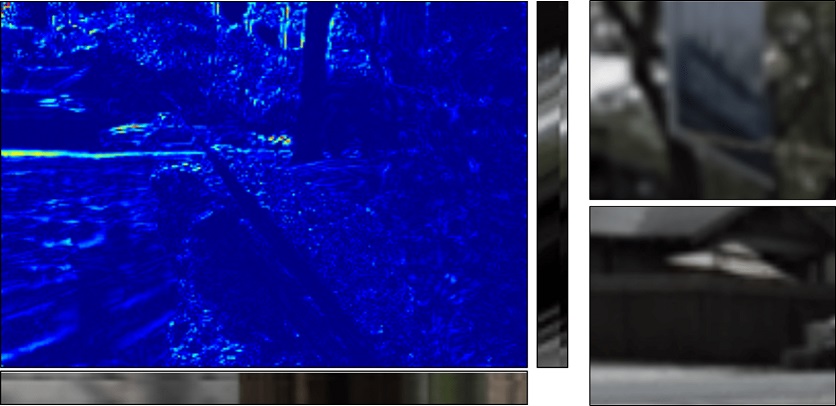}}~%
    
    \vspace{1.0mm}
    {\includegraphics[width=0.33\linewidth]{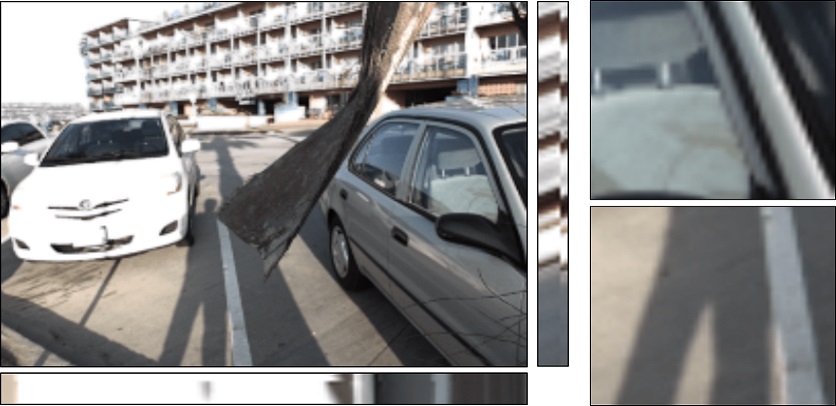}}~%
    {\includegraphics[width=0.33\linewidth]{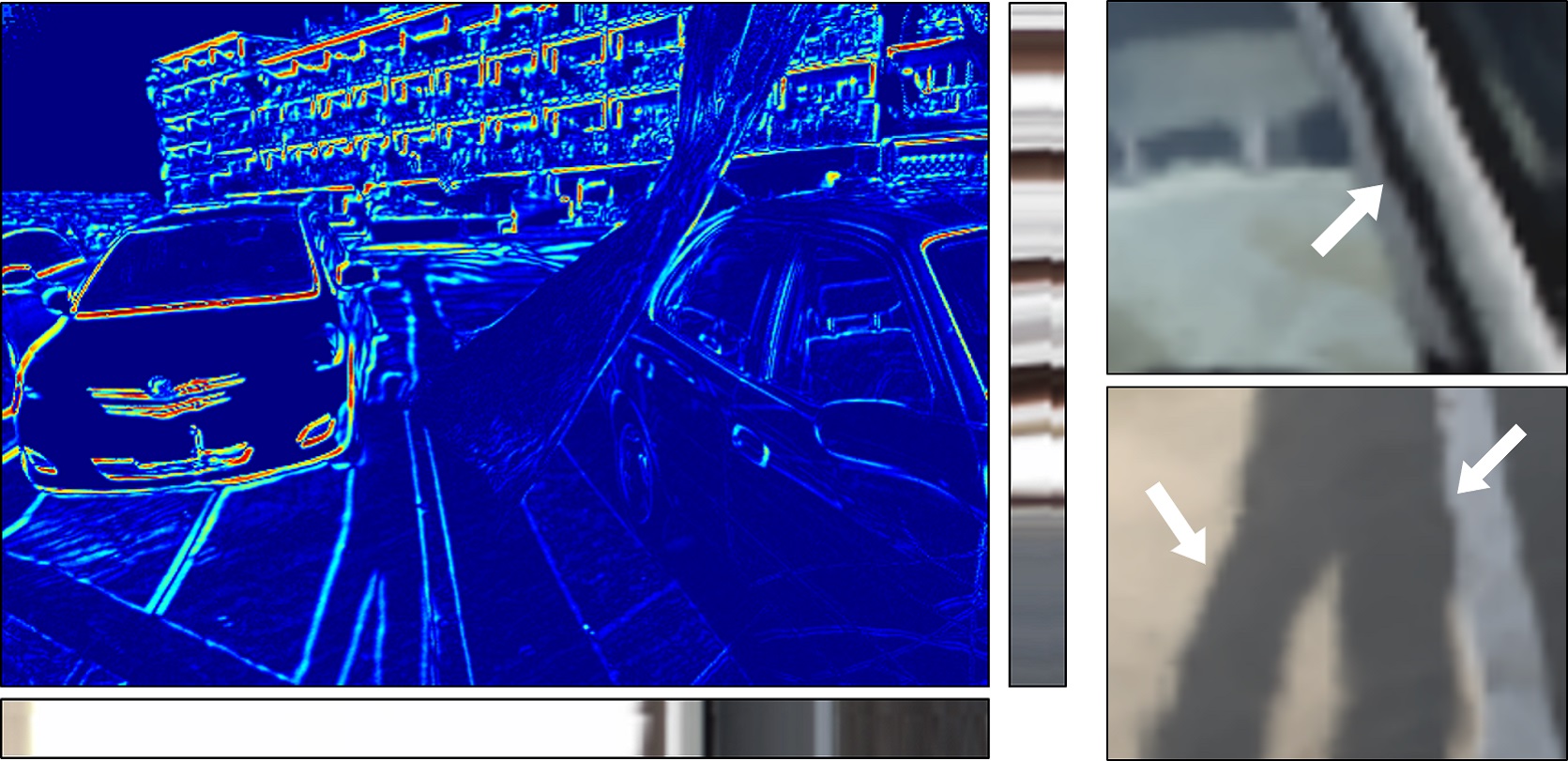}}~%
    {\includegraphics[width=0.33\linewidth]{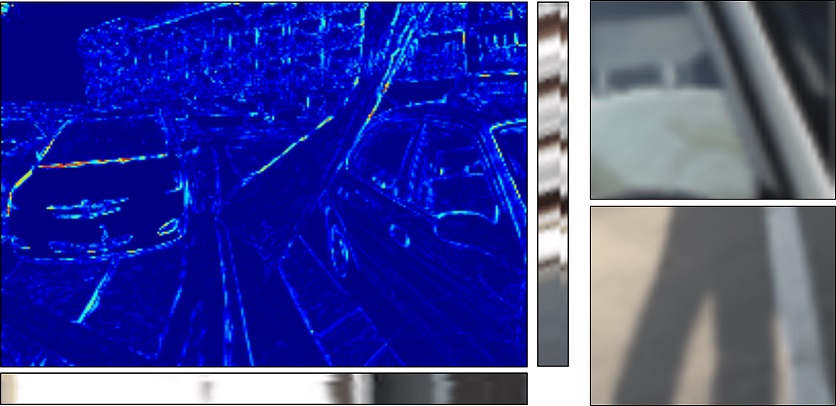}}~%
    
    \vspace{1.0mm}
    {\includegraphics[width=0.33\linewidth]{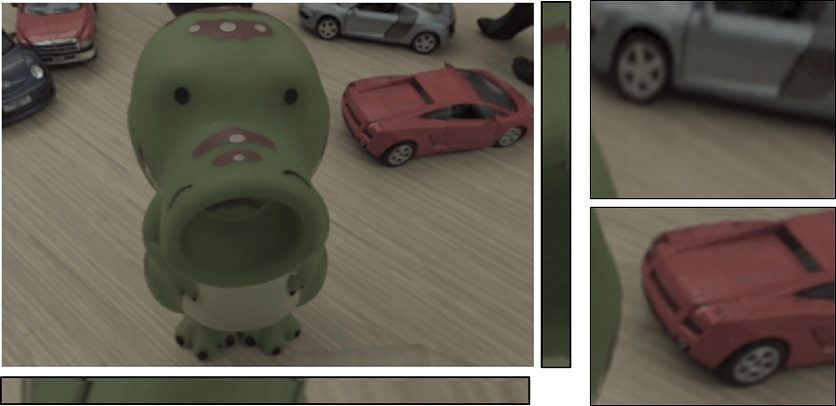}}~%
    {\includegraphics[width=0.33\linewidth]{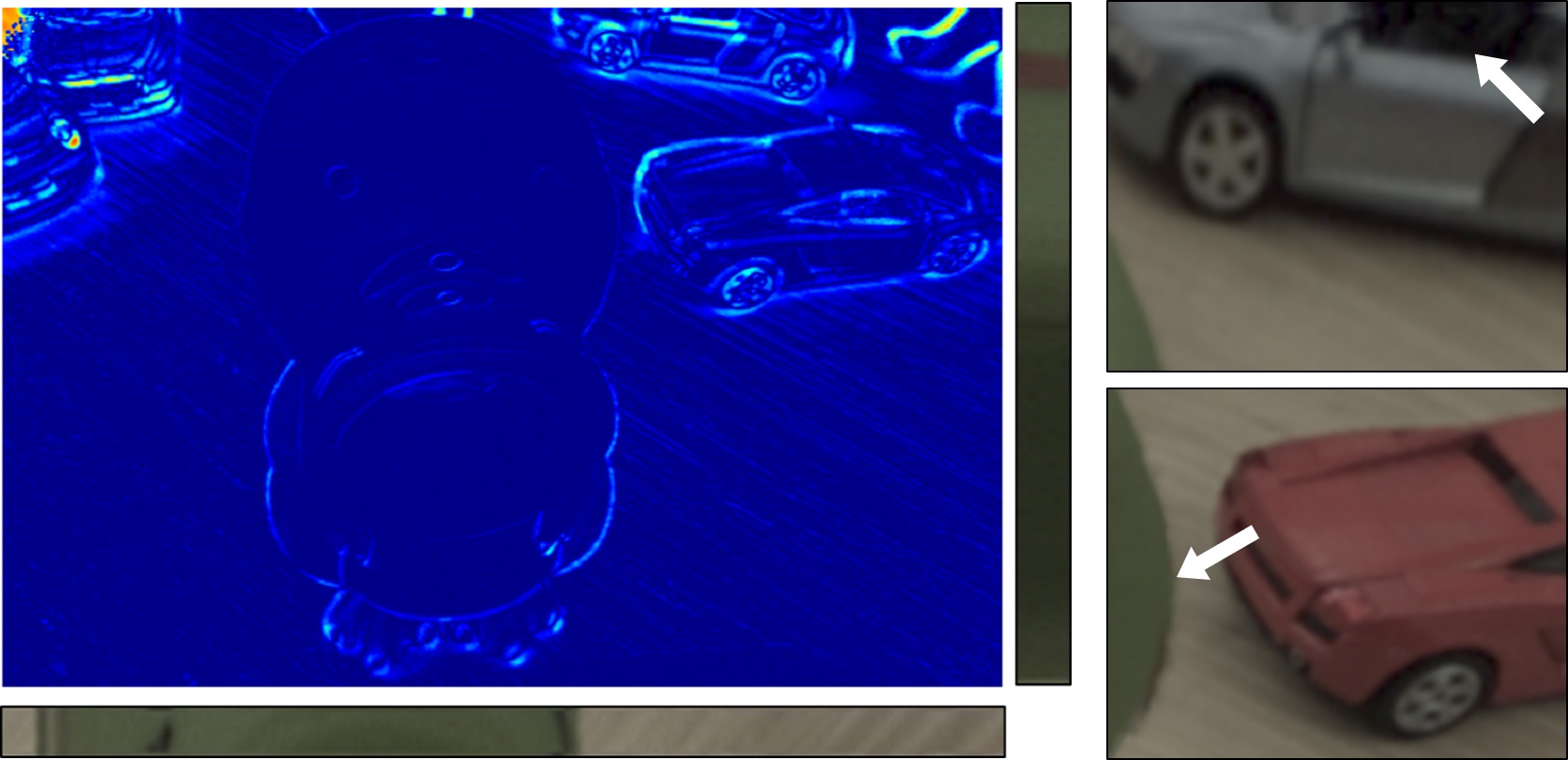}}~%
    {\includegraphics[width=0.33\linewidth]{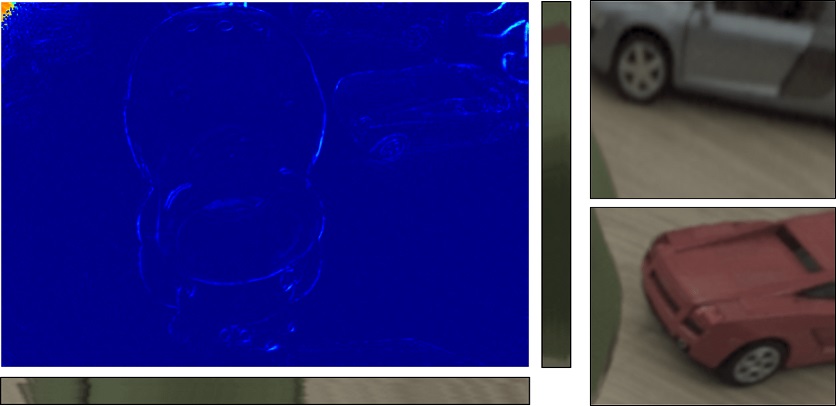}}~%
    
    \vspace{1.0mm}
	\stackunder[5pt]{\includegraphics[width=0.33\linewidth]{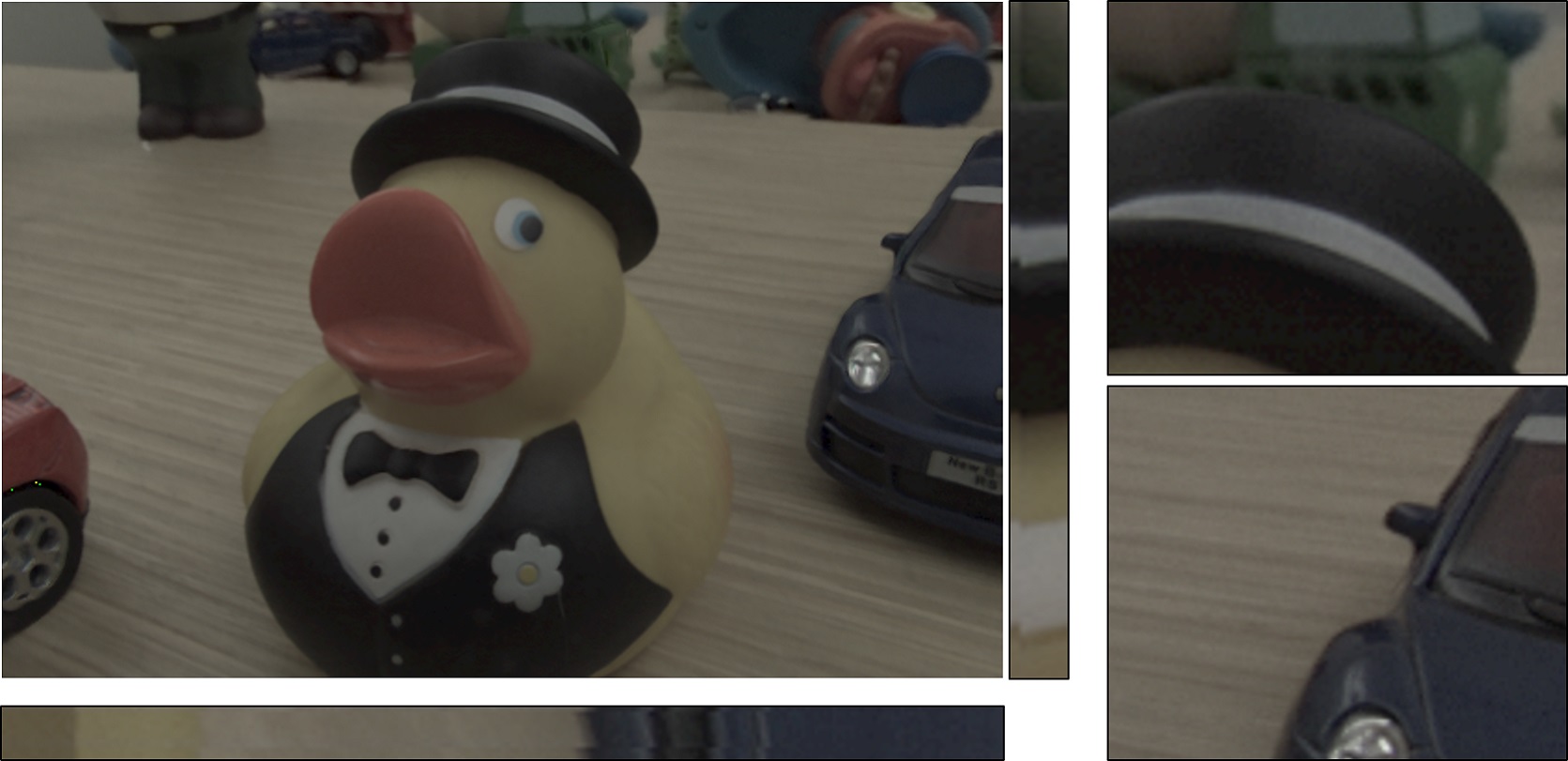}}{Reference image}~%
	\stackunder[5pt]{\includegraphics[width=0.33\linewidth]{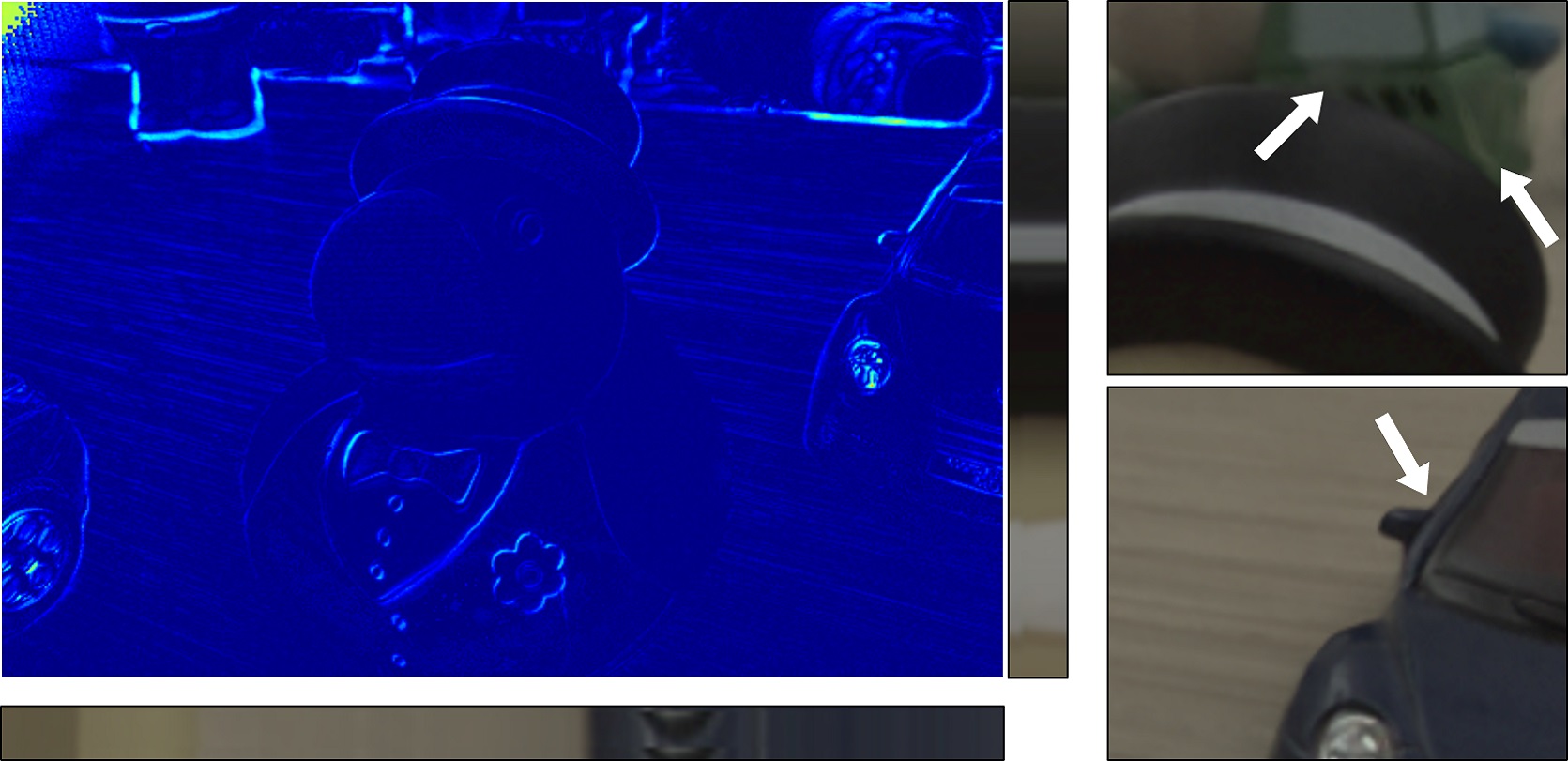}}{Srininvasan~\etal~\cite{Srinivasan_ICCV17}}~%
	\stackunder[5pt]{\includegraphics[width=0.33\linewidth]{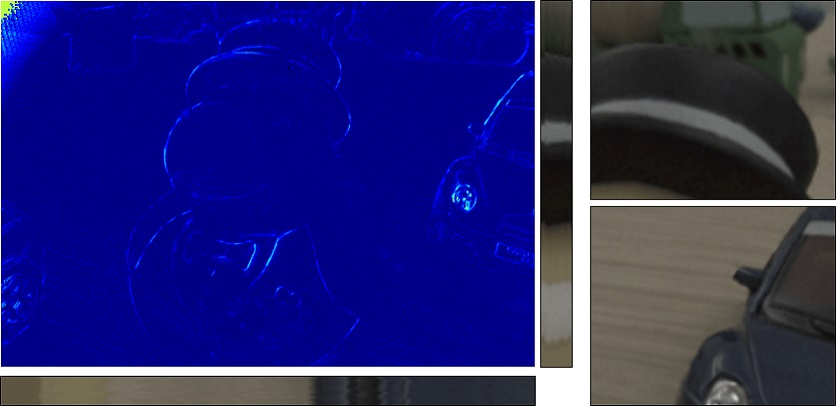}}{Proposed}~%

\end{center}
\vspace{-2.5mm}
\caption{Qualitative evaluation on \textit{Flower}, \textit{Toys}, and \textit{General} dataset. EPIs shown are scaled and cropped for better visibility. The error map and sliced EPIs from our method indicates geometrically better light field across all datasets. From the top to bottom are Flower~\#5, Flower~\#10, Flower~\#12, Rock, Cars, IMG\textunderscore 1827, and IMG\textunderscore 9998.}
    \label{fig:Main_Compare}
\end{figure*}
%------------------------------------------------------------------------
\section{Experimental Results}

We evaluate the performance of the proposed framework qualitatively and quantitatively.
We then compare its performance with that of a state-of-the-art method of light field synthesis from a single image ~\cite{Srinivasan_ICCV17}.

We utilize three datasets, namely, \textit{Flower}, \textit{Toys}~\cite{Srinivasan_ICCV17}, and \textit{General}~\cite{Kalantari_TOG16}.
For the evaluation, all networks are re-trained with the corresponding dataset.
The network is trained in an end-to-end fashion for 150,000 iterations with a batch size of 1.
Hyperparameters ~$\eta$, $\lambda_e$, $\lambda_g$ and $\lambda_{tv}$ are set to 0.8, 10, 10 and $1e^{-6}$, respectively.
We use Adam optimizer~\cite{Kingma_ICLR14} as our optimization algorithm with the default parameters.
The proposed network is implemented using TensorFlow~\cite{Abadi_OSDI16}.
Training is performed for approximately 15 h on NVIDIA GTX 1080Ti GPU with 11GB of memory and Intel i7-7700 @3.60 GHz CPU with 16GB of memory.
For the inference time, the network runs approximately for 0.5 s when synthesizing a single light field image.
Readers are urged to watch the supplementary video for improved understanding and elaborated results.
More technical details regarding training are available in supplementary material.
%-------------------------------------------------------------------------
\begin{table}
\resizebox{\linewidth}{!}{%
\begin{tabular}{@{}ccccccc@{}}
\toprule
Dataset                                                     & \multicolumn{2}{c}{Flower}     & \multicolumn{2}{c}{Toys}       & \multicolumn{2}{c}{General}    \\ \midrule
Metric                                                     & PSNR           & SSIM          & PSNR           & SSIM          & PSNR           & SSIM          \\ \midrule
Kalantari~\cite{Kalantari_TOG16}    & -              & -             & -              & -             & \textbf{37.50*} & \textbf{0.97*} \\
Srinivasan~\cite{Srinivasan_ICCV17} & 31.69          & 0.85          & 33.89          & 0.83          & 28.89          & 0.86          \\
Proposed                                                   & \textbf{36.31} & \textbf{0.93} & \textbf{36.43} & \textbf{0.87} & 29.82          & 0.86          \\ \bottomrule
\end{tabular}%
}
\caption{Average PSNR (in dB) and SSIM from \textit{Flower} (200 images), \textit{Toys} (100 images) and \textit{General} (30 images) datasets.}
  \label{table:Quantitative_General}
\vspace{-2.5mm}
\end{table}
%------------------------------------------------------------------------
%------------------------------------------------------------------------
\subsection{\textit{\textbf{Flower}} Dataset}
The light field images in the \textit{Flower} dataset have a clear distinction between the background and foreground.
The flower in the foreground has a more dominant color than the object in the background.
In our work, we carefully randomize our test set (200 images) so as not to contain any data in the training set (3,100 images).
We remove unusable light field images that contains considerable noise.
To train and test the dataset, we downsample the original size into 256$\times$192$\times$8$\times$8 by using bilinear interpolation due to the GPU memory limitation.
To evaluate performance, we upsample the synthesized light field into the original size before measuring the PSNR and SSIM metrics.
Therefore, a fair comparison between the state-of-the-art method~\cite{Srinivasan_ICCV17} is conducted at the original size.

Figure~\ref{fig:Quantitative_Flower} shows a comparison between the proposed method and Srinivasan's work~\cite{Srinivasan_ICCV17} using 20 random samples.
The average PSNR and SSIM for the \textit{Flower} test set is shown in Table~\ref{table:Quantitative_General}.
The proposed method significantly outperforms the state-of-the-art method significantly in most cases.
Moreover, the average PSNR and SSIM metric increases by approximately 4~dB and 0.07, respectively.

Figure~\ref{fig:Main_Compare} shows a qualitative evaluation of data~\#5, data~\#10, and \#12.
Data~\#5 and ~\#10 have multiple flowers with a similar color at different locations.
Srinivasan's work fails on this kind of scene due to its dependency on finding a single object with a dominant color, which leads to inaccurate depth estimation.
As shown in the error map, pixels with the flower in the back have many errors.
Data~\#12 shows scene with single flower commonly found in the \textit{Flower} dataset.
The error map and zoomed in images shows the proposed method handles occlusion better and have less artifact.
Meanwhile, our network is capable of robustly estimating the flower geometry and location.
Data~\#10 is an example of difficult data with specular in the background.
Srinivasan's work cannot estimate depth in those regions.
Overall, the proposed work performs better than Srinivasan's method due to our image shifting approach and the proposed loss function.

\subsection{\textit{\textbf{Toys}} and \textit{\textbf{General}} Dataset}
We utilize \textit{Toys} and \textit{General} datasets to verify the performance of the proposed model.
Unlike the \textit{Flower} dataset, \textit{Toys} and \textit{General} datasets have more variance in object class in arbitrary location.
Figure~\ref{fig:Main_Compare} shows the qualitative evaluation of the two datasets.
Cars and IMG\textunderscore 1827 examples show similar limitation of Srinivasan~\cite{Srinivasan_ICCV17} in the \textit{Flower} dataset.
Furthermore, Srinivasan~\cite{Srinivasan_ICCV17} results on IMG\textunderscore 1827 shows incorrect EPI slope.
The slope direction is inverted indicating the shifting direction of the object is incorrect.
While the proposed method successfully synthesizes a proper light field image due to the power of proposed light field based loss function.

The quantitative result is shown in Table~\ref{table:Quantitative_General}.
Kalantari's work~\cite{Kalantari_TOG16} is expected to outperform the proposed work because the former uses four inputs to synthesize the light field.
It only serves as the upper baseline of this specific dataset and not for any comparison due to the different number of input images.
However, the proposed network outperforms~\cite{Srinivasan_ICCV17}, which proves its capability for general scenes.

%------------------------------------------------------------------------
\begin{table}[]
\begin{tabular}{@{}ccccc@{}}
\toprule
\multirow{2}{*}{Method} & \multicolumn{2}{c}{Flower} & \multicolumn{2}{c}{General} \\ \cmidrule(l){2-5}
                        & PSNR         & SSIM        & PSNR          & SSIM        \\ \midrule
without post-processing & 36.31        & 0.931       & 29.82         & 0.86        \\
with post-processing    & 36.55        & 0.932       & 30.92         & 0.87        \\ \bottomrule
\end{tabular}
\caption{Quantitative improvement of post-processing.}
  \label{table:Post_Processing}
\vspace{-2.5mm}
\end{table}
%------------------------------------------------------------------------
%------------------------------------------------------------------------
\begin{table}[]
\centering
\resizebox{1.0\linewidth}{!}
{%
\begin{tabular}{@{}cccccc@{}}
\toprule
Metric & L1    & Global & Local & L+G   & w/o Image Shifting \\ \midrule
PSNR   & 28.14 & 30.52  & 35.02 & \textbf{36.31} & 33.73                \\
SSIM   & 0.77  & 0.87   & 0.92  & \textbf{0.93}  & 0.90                \\ \bottomrule
\end{tabular}%
}
\caption{Quantitative evaluation of each loss function's effect to the network trained using \textit{Flower} dataset.}
  \label{table:Loss_Effect}
\vspace{-2.5mm}
\end{table}
%------------------------------------------------------------------------
\subsection{Post-processing}
Flow extraction from a complex scene is difficult, especially in the discontinuity region.
Incorrect estimation leads to artifacts that reduce the light field coherence in the angular domain.
To handle this problem, we propose a post-processing method inspired by cost aggregation in stereo matching.
We regard appearance flow as the cost volume and apply an edge-aware filter (guided filter~\cite{He_PAMI13}) to it.
Through the flow aggregation, noisy flow is smoothed, and important edge information are maintained.
Flow aggregation is performed before applying the bilinear sampler module.
Post-processing improves the performance by up to 1 dB for the \textit{General} dataset.
As shown in Table~\ref{table:Post_Processing}, post-processing does not considerably affect the \textit{Flower} dataset because, the estimated appearance flow is already accurate.
%-------------------------------------------------------------------------
\begin{figure}[t]
\begin{center}
	\subfloat[Reference image]
	{	
	{\includegraphics[width=0.47\linewidth]{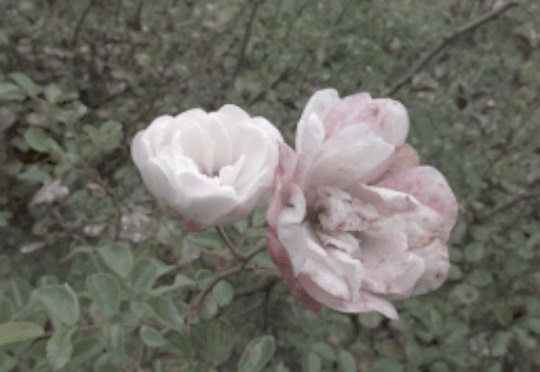}}~%
    {\includegraphics[width=0.47\linewidth]{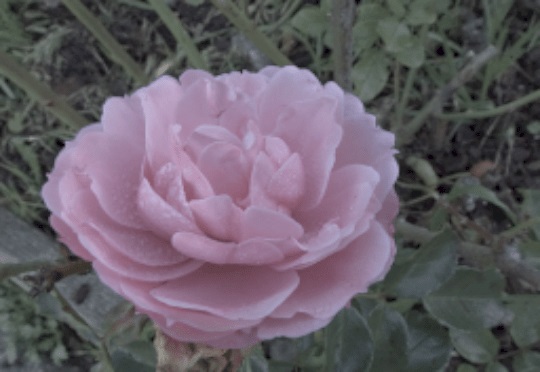}}~%	
	}
	
    \vspace{-2.5mm}
    \subfloat[Depth from synthesized LF and refocus image at foreground]
	{
	{\includegraphics[width=0.47\linewidth]{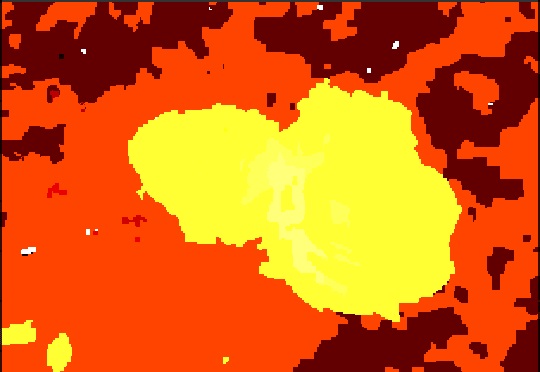}}~%
    {\includegraphics[width=0.47\linewidth]{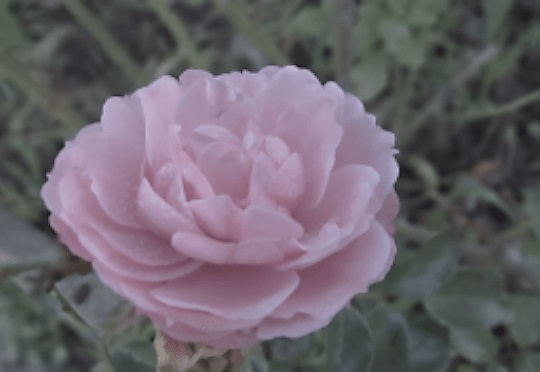}}~%
    }
    
    \vspace{-2.5mm}
	\subfloat[Depth from ground truth LF and refocus image at background]
	{
	{\includegraphics[width=0.47\linewidth]{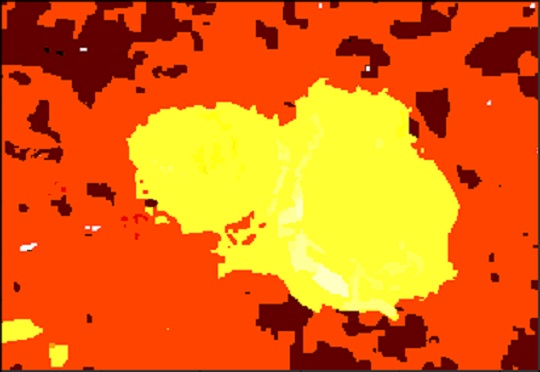}}~%
    {\includegraphics[width=0.47\linewidth]{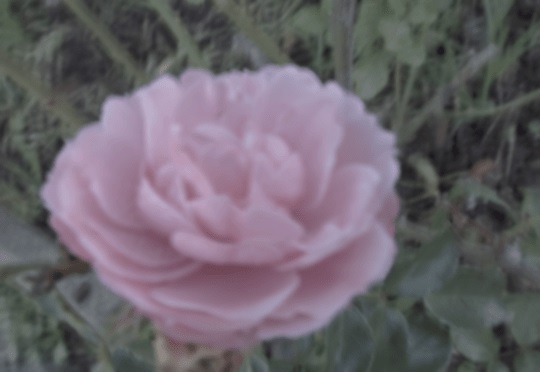}}~%
    }
	\vspace{-3.5mm}
\end{center}
\caption{Visual comparison of depth estimation using CAE~\cite{Williem_PAMI17} and synthetic defocus result.}
    \label{fig:Depth_Refocus_Result}
\vspace{-2.0mm}
\end{figure}
%-------------------------------------------------------------------------
\subsection{Ablation Study}
To reveal the discriminate power of the proposed light field synthesis network, we evaluate the performance of different loss functions, namely, pixel-wise L1 loss, global loss, local loss, and local--global loss.
Table~\ref{table:Loss_Effect} shows the quantitative evaluation of each loss function's effect on the network.
The simple pixel-wise loss is outperformed by the proposed loss functions.
Meanwhile, individually, local loss achieves satisfactory performance, and the combination with global loss leads to improved performance.
Global and local losses complement each other by forcing local and global consistency.
Additionally, we also show that the synthesized light field can be used for common light field applications, such as depth estimation and refocusing.
Figure~\ref{fig:Depth_Refocus_Result} shows convincing depth estimation and synthetic defocus blur result.
This findings confirms that the synthesized light field is geometrically correct and can be used in various light field processing applications.

\section{Conclusion}
In this study, we develop an end-to-end deep model for light field synthesis from a single image by using appearance flow.
Novel light field based loss functions are introduced to preserve spatio-angular consistency and remove the dependency of pixel intensity.
A regularization and post-processing strategy are presented to address complex scene.
The experimental results show that the proposed network qualitatively and quantitatively  outperforms the state-of-the-art single image based light field synthesis algorithm.
The proposed method can also be generalized to arbitrary scenes better than previous work.
Future work will include wide-baseline light field synthesis and explicit handling of view-dependent reflection.

{\small
\bibliographystyle{ieee}
\bibliography{egbib}

\begin{thebibliography}{10}\itemsep=-1pt

\bibitem{Raytrix_2013}
{Raytrix} {3D} light field camera.
\newblock \url{https://raytrix.de/products/}.
\newblock Accessed: 2018-09-08.

\bibitem{Abadi_OSDI16}
M.~Abadi, P.~Barham, J.~Chen, Z.~Chen, A.~Davis, J.~Dean, M.~Devin,
  S.~Ghemawat, G.~Irving, M.~Isard, et~al.
\newblock Tensorflow: a system for large-scale machine learning.
\newblock In {\em OSDI}, volume~16, pages 265--283, 2016.

\bibitem{Dong_PAMI16}
C.~Dong, C.~C. Loy, K.~He, and X.~Tang.
\newblock Image super-resolution using deep convolutional networks.
\newblock {\em IEEE Trans. on Pattern Analysis and Machine Intelligence},
  38(2):295--307, 2016.

\bibitem{Flynn_CVPR16}
J.~Flynn, I.~Neulander, J.~Philbin, and N.~Snavely.
\newblock Deepstereo: Learning to predict new views from the world's imagery.
\newblock In {\em Proc. of IEEE Conference on Computer Vision and Pattern
  Recognition}, pages 5515--5524, 2016.

\bibitem{Garg_ECCV16}
R.~Garg, V.~K. BG, G.~Carneiro, and I.~Reid.
\newblock Unsupervised cnn for single view depth estimation: Geometry to the
  rescue.
\newblock In {\em Proc. of European Conference on Computer Vision}, pages
  740--756, 2016.

\bibitem{Godard_CVPR17}
C.~Godard, O.~Mac~Aodha, and G.~J. Brostow.
\newblock Unsupervised monocular depth estimation with left-right consistency.
\newblock In {\em Proc. of IEEE Conference on Computer Vision and Pattern
  Recognition}, volume~2, page~7, 2017.

\bibitem{He_PAMI13}
K.~He, J.~Sun, and X.~Tang.
\newblock Guided image filtering.
\newblock {\em IEEE Trans. on Pattern Analysis and Machine Intelligence},
  35(6):1397--1409, 2013.

\bibitem{Jaderberg_NIPS15}
M.~Jaderberg, K.~Simonyan, A.~Zisserman, et~al.
\newblock Spatial transformer networks.
\newblock In {\em Proc. of Advances in neural information processing systems},
  pages 2017--2025, 2015.

\bibitem{Johannsen_ACCV16}
O.~Johannsen, A.~Sulc, N.~Marniok, and B.~Goldluecke.
\newblock Layered scene reconstruction from multiple light field camera views.
\newblock In {\em Proc. of Asian Conference on Computer Vision}, pages 3--18,
  2016.

\bibitem{Kalantari_TOG16}
N.~K. Kalantari, T.-C. Wang, and R.~Ramamoorthi.
\newblock Learning-based view synthesis for light field cameras.
\newblock {\em ACM Trans. on Graphics}, 35(6):193, 2016.

\bibitem{Kingma_ICLR14}
D.~P. Kingma and J.~Ba.
\newblock Adam: A method for stochastic optimization.
\newblock In {\em Proc. of International Conference on Machine Learning}, 2015.

\bibitem{Levoy_CGI96}
M.~Levoy and P.~Hanrahan.
\newblock Light field rendering.
\newblock In {\em Proc. of SIGGRAPH}, pages 31--42, 1996.

\bibitem{Nair_ICML10}
V.~Nair and G.~E. Hinton.
\newblock Rectified linear units improve restricted boltzmann machines.
\newblock In {\em Proc. of International Conference on Machine Learning}, pages
  807--814, 2010.

\bibitem{Ng_CSTR05}
R.~Ng, M.~Levoy, M.~Br{\'e}dif, G.~Duval, M.~Horowitz, P.~Hanrahan, et~al.
\newblock Light field photography with a hand-held plenoptic camera.
\newblock {\em Computer Science Technical Report CSTR}, 2(11):1--11, 2005.

\bibitem{Ruan_EG18}
L.~Ruan, B.~Chen, and M.~L. Lam.
\newblock {Light Field Synthesis from a Single Image using Improved Wasserstein
  Generative Adversarial Network}.
\newblock In {\em Eurographics - Posters}, 2018.

\bibitem{Schilling_CVPR18}
H.~Schilling, M.~Diebold, C.~Rother, and B.~J{\"a}hne.
\newblock Trust your model: Light field depth estimation with inline occlusion
  handling.
\newblock In {\em Proc. of IEEE Conference on Computer Vision and Pattern
  Recognition}, pages 4530--4538, 2018.

\bibitem{Shin_CVPR18}
C.~Shin, H.-G. Jeon, Y.~Yoon, I.~S. Kweon, and S.~J. Kim.
\newblock Epinet: A fully-convolutional neural network using epipolar geometry
  for depth from light field images.
\newblock In {\em Proc. of IEEE Conference on Computer Vision and Pattern
  Recognition}, pages 4748--4757, 2018.

\bibitem{Srinivasan_ICCV17}
P.~P. Srinivasan et~al.
\newblock Learning to synthesize a 4{D} {RGBD} light field from a single image.
\newblock In {\em Proc. of IEEE International Conference on Computer Vision},
  2017.

\bibitem{Tao_ICCV13}
M.~W. Tao, S.~Hadap, J.~Malik, and R.~Ramamoorthi.
\newblock Depth from combining defocus and correspondence using light-field
  cameras.
\newblock In {\em Proc. of IEEE International Conference on Computer Vision},
  pages 673--680, 2013.

\bibitem{Vianello_CVPR18}
A.~Vianello, J.~Ackermann, R.~B. Campus, M.~Diebold, and B.~J{\"a}hne.
\newblock Robust hough transform based {3D} reconstruction from circular light
  fields.
\newblock In {\em Proc. of IEEE Conference on Computer Vision and Pattern
  Recognition}, pages 7327--7335, 2018.

\bibitem{Wang_ECCV18}
Y.~Wang, F.~Liu, Z.~Wang, G.~Hou, Z.~Sun, and T.~Tan.
\newblock End-to-end view synthesis for light field imaging with pseudo
  {4DCNN}.
\newblock In {\em Proc. of European Conference on Computer Vision}, pages
  340--355, 2018.

\bibitem{Wanner_PAMI14}
S.~Wanner and B.~Goldluecke.
\newblock Variational light field analysis for disparity estimation and
  super-resolution.
\newblock {\em IEEE Trans. on Pattern Analysis and Machine Intelligence},
  36(3):606--619, 2014.

\bibitem{Wilburn_TOG05}
B.~Wilburn, N.~Joshi, V.~Vaish, E.-V. Talvala, E.~Antunez, A.~Barth, A.~Adams,
  M.~Horowitz, and M.~Levoy.
\newblock High performance imaging using large camera arrays.
\newblock In {\em ACM Trans. on Graphics}, volume~24, pages 765--776, 2005.

\bibitem{Williem_CVPR16}
Williem and I.~K. Park.
\newblock Robust light field depth estimation for noisy scene with occlusion.
\newblock In {\em Proc. of IEEE Conference on Computer Vision and Pattern
  Recognition}, pages 4396--4404, 2016.

\bibitem{Williem_PAMI17}
Williem, I.~K. Park, and K.~M. Lee.
\newblock Robust light field depth estimation using occlusion-noise aware data
  costs.
\newblock {\em IEEE Trans. on Pattern Analysis and Machine Intelligence},
  (10):2484--2497, 2018.

\bibitem{Wu_CVPR17}
G.~Wu, M.~Zhao, L.~Wang, Q.~Dai, T.~Chai, and Y.~Liu.
\newblock Light field reconstruction using deep convolutional network on epi.
\newblock In {\em Proc. of IEEE Conference on Computer Vision and Pattern
  Recognition}, volume 2017, page~2, 2017.

\bibitem{Xie_ECCV16}
J.~Xie, R.~Girshick, and A.~Farhadi.
\newblock Deep{3D}: Fully automatic {2D}-to-{3D} video conversion with deep
  convolutional neural networks.
\newblock In {\em Proc. of European Conference on Computer Vision}, pages
  842--857, 2016.

\bibitem{Yeung_ECCV18}
H.~W.~F. Yeung, J.~Hou, J.~Chen, Y.~Y. Chung, and X.~Chen.
\newblock Fast light field reconstruction with deep coarse-to-fine modelling of
  spatial-angular clues.
\newblock In {\em Proc. of European Conference on Computer Vision}, 2018.

\bibitem{Zhang_CVPR15}
Z.~Zhang, Y.~Liu, and Q.~Dai.
\newblock Light field from micro-baseline image pair.
\newblock In {\em Proc. of IEEE Conference on Computer Vision and Pattern
  Recognition}, pages 3800--3809, 2015.

\bibitem{Zhou_ECCV16}
T.~Zhou et~al.
\newblock View synthesis by appearance flow.
\newblock In {\em Proc. of ECCV}, 2016.

\bibitem{Zhou_TOG18}
T.~Zhou, R.~Tucker, J.~Flynn, G.~Fyffe, and N.~Snavely.
\newblock Stereo magnification: Learning view synthesis using multiplane
  images.
\newblock {\em ACM Trans. on Graphics}, 37(4):65:1--65:12, 2018.

\end{thebibliography}
}

\end{document}